%% file: main.tex
\documentclass[10pt,journal,twoside]{IEEEtran}
\usepackage{amsmath}
\usepackage{graphicx}

\usepackage{multirow}

\usepackage{tabu}
\usepackage{subfigure}
\usepackage{amsfonts}
\usepackage{amssymb,cool}
\usepackage{multirow}
\usepackage{booktabs} 
\usepackage{stfloats}
\usepackage[normalem]{ulem}
\usepackage{cite}

\DeclareMathOperator*{\argmin}{argmin}
\usepackage{algorithm}
\usepackage{algorithmic}
\usepackage{color}
\usepackage{url}

\usepackage{url}

\usepackage[capitalise]{cleveref}

\ifCLASSINFOpdf
\else
\fi
\begin{document}
%

\title{Multitask Identity-Aware Image Steganography via Minimax Optimization}
%
%
%

\author{Jiabao~Cui,
        Pengyi~Zhang,
        Songyuan~Li,
        Liangli~Zheng,
        Cuizhu~Bao,
        Jupeng~Xia,
        and~Xi~Li
        \thanks{Manuscript received October 19, 2020; revised May 16, 2021. }
        \thanks{J.~Cui, P. Zhang, S.~Li, and X.~Li are with the College of Computer Science and Technology, Zhejiang University, Hangzhou 310027, China. E-mail: \{jbcui, pyzhang, leizungjyun, xilizju\}@zju.edu.cn.}
        \thanks{L. Zheng is with the School of Software Technology, Zhejiang University, Ningbo 315048, China. E-mail: lianglizheng@zju.edu.cn.}
        \thanks{C.~Bao is with the College of Computer Science and Information Engineering, Zhejiang Gongshang University, Hangzhou 310018, China. E-mail: baocuizhu@zjgsu.edu.cn.}
        \thanks{J.~Xia is with the Department of Security Technology, Ant Group, Hangzhou 310000 China. E-mail: jupeng.xia@antgroup.com.}
\thanks{(Corresponding author: Xi Li.)}
}




%
%

\markboth{IEEE TRANSACTIONS ON IMAGE PROCESSING,~Vol.~xx, No.~x, July~2020}%
{Cui \MakeLowercase{\textit{et al.}}: Multitask Identity-Aware Image Steganography via Minimax Optimization}
%



\maketitle


\input{sections/abstract.tex}


%
\IEEEpeerreviewmaketitle

\input{sections/intro.tex}

\input{sections/related.tex}

\input{sections/method.tex}

\input{sections/experiment.tex}

\input{sections/conclusion.tex}

\ifCLASSOPTIONcaptionsoff
  \newpage
\fi



%
%
%

\bibliographystyle{IEEEtran}
\bibliography{IEEEabrv,sample-base}

%

%
%
%



\begin{IEEEbiography}[{\includegraphics[width=1in,height=1.25in,clip,keepaspectratio]{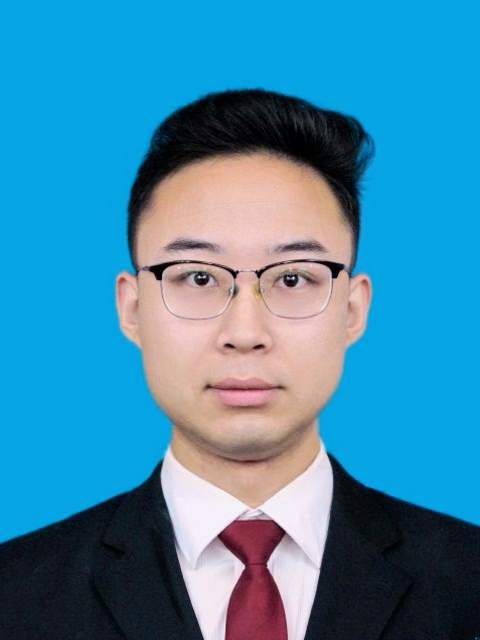}}]{Jiabao Cui}
	received the B.S. degree in computer science and technology from Xidian University, Xi'an, ShaanXi, China. He is currently studying for the PhD degree at the College of Computer Science at Zhejiang University, Hangzhou, China. His research interests include computer vision, image classification, image steganography and machine learning.
\end{IEEEbiography}
\vspace{1em}

\begin{IEEEbiography}[{\includegraphics[width=1in,height=1.25in,clip,keepaspectratio]{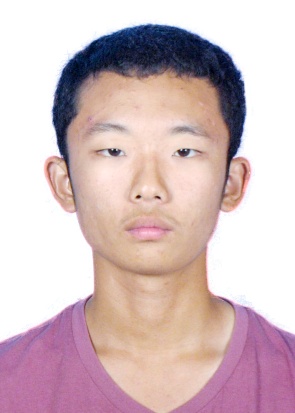}}]{Pengyi Zhang}
	received the bachelor's degree in computer science and technology from Beijing Institute of Technology, China, in 2020. He is currently pursuing the master's degree with the College of Computer Science, Zhejiang University, Hangzhou, China, under the supervision of Prof. X. Li. His current research interests are primarily in computer vision and deep learning.
\end{IEEEbiography}
\vspace{1em}

\begin{IEEEbiography}[{\includegraphics[width=1in,height=1.25in,clip,keepaspectratio]{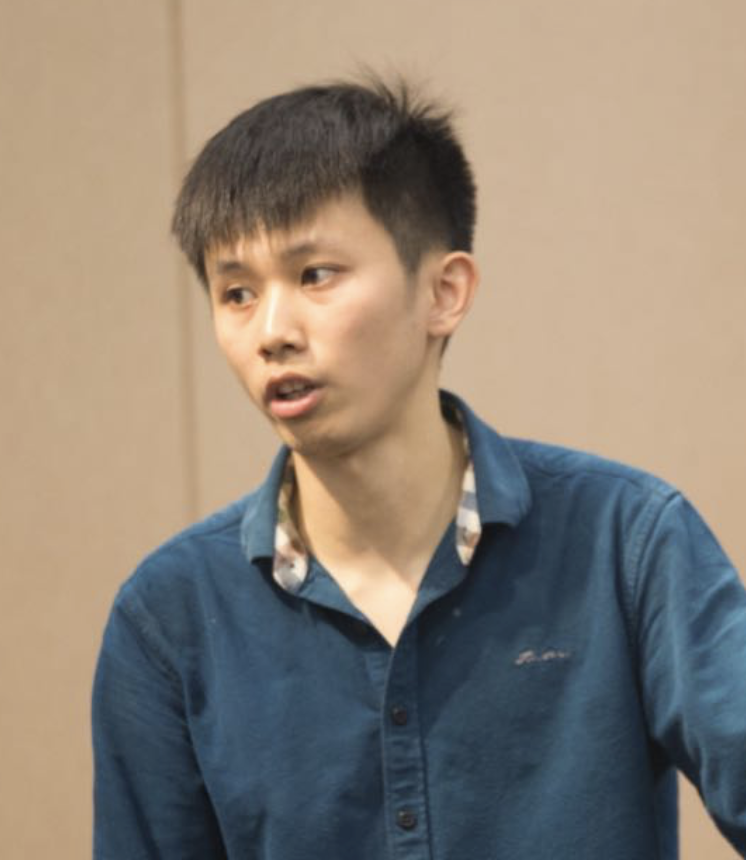}}]{Songyuan Li}
	received his master's degree in 2017 from Zhejiang University, China, where he worked on problems in computer architecture and operating systems. He is currently a Ph.D. candidate at Zhejiang University. His current research interests include knowledge distillation, semantic segmentation and dynamic routing.
\end{IEEEbiography}
\vspace{1em}

\begin{IEEEbiography}[{\includegraphics[width=1in,height=1.25in,clip,keepaspectratio]{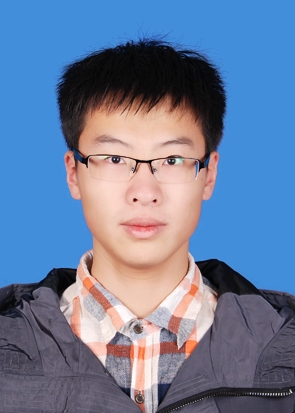}}]{Liangli Zheng}
	received the bachelor's degree in information technology from Nanjing University of Information Science and Technology, China, in 2020. He is currently pursuing the master's degree with the College of Software Technology, Zhejiang University, Hangzhou, China, under the supervision of Prof. X. Li. His current research interests are primarily in computer vision and deep learning.
\end{IEEEbiography}
\vspace{1em}

\begin{IEEEbiography}[{\includegraphics[width=1in,height=1.25in,clip,keepaspectratio]{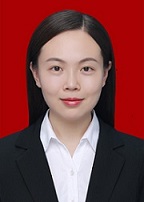}}]{Cuizhu Bao} 
	received the doctor degree in traffic information engineering and control from Jilin University, Chang’chun, Jilin, China. She is currently teaching at the College of Computer Science and Information Engineering at Zhejiang Gongshang University, Hangzhou, China. Her research interests include computer vision, image classification, machine learning and Visual relationship detection.
\end{IEEEbiography}
\vspace{1em}

\begin{IEEEbiography}[{\includegraphics[width=1in,height=1.25in,clip,keepaspectratio]{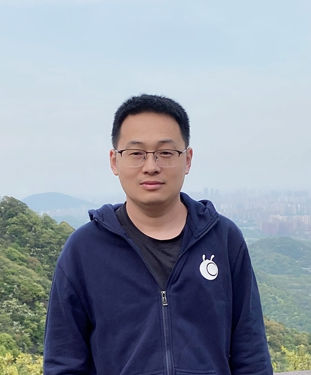}}]{Jupeng Xia}
	has worked with Security Technology Department of Ant Group since
	2010, where he is currently a Team Leader of the Data Security and Privacy Protection, experienced in endpoint security, network security, cryptography and data intelligence for risk \& compliance. He is currently in charge of research \& development for data security at Ant Group.
\end{IEEEbiography}
\vspace{1em}

\begin{IEEEbiography}[{\includegraphics[width=1in,height=1.25in,clip,keepaspectratio]{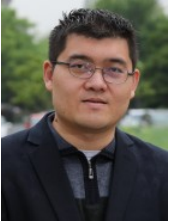}}]{Xi $\textup{Li}^{\dagger}$}
	received the Ph.D. degree from the National Laboratory of Pattern Recognition, Chinese Academy of Sciences, Beijing, China, in 2009. From 2009 to 2010, he was a Post-Doctoral Researcher with CNRS Telecom ParisTech, France. He was a Senior Researcher with the University of Adelaide, Australia. He is currently a Full Professor with Zhejiang University, China. His research interests include visual tracking, compact learning, motion analysis, face recognition, data mining, and image retrieval.
\end{IEEEbiography}

\end{document}


%
\title{Supplementary Material:\\
	\huge{``Multitask Identity-Aware Image Steganography via Minimax Optimization"}}
\author{
	Jiabao Cui, Pengyi Zhang, Songyuan Li, Liangli Zheng, Cuizhu Bao, Jupeng~Xia, and Xi~Li
}

\IEEEtitleabstractindextext{
\begin{abstract}
In this supplementary material, we first state the coefficients among the loss terms in the experiments (Section \uppercase\expandafter {\romannumeral 3}-C of the main text). Secondly, we investigate the hiding threshold and explain the rationality of our hiding threshold setting (Section~\uppercase\expandafter {\romannumeral 4}-A of the main text). Thirdly, we show the visual quality of different ablation experiments and all the competing methods of visual information hiding (Section~\uppercase\expandafter {\romannumeral 4}-B of the main text). Finally, we show the visual quality of different image steganography methods (Section~\uppercase\expandafter {\romannumeral 4}-C of the main text).
\end{abstract}
}
\maketitle
\renewcommand{\theequation}{S-\arabic{equation}}
\setcounter{equation}{0}
\renewcommand{\thefigure}{S-\arabic{figure}}
\setcounter{figure}{0}
\renewcommand{\thetable}{S-\arabic{table}}
\setcounter{table}{0}

\section{Coefficients in the Experiments} \label{apdx_norm_iterm}

%
In this section, we specify the coefficients of Eqs.~(12), (13), and (14) on the experiments on recognition and multi-task learning for reproduction, as shown in \mbox{\cref{coeffients}}.

\input{../tables/coeffients.tex}

%

\section{Hiding Threshold Selection} \label{apdx_Hiding_threshold}

\begin{figure}[h]
	\centering
	\includegraphics[width=.45\textheight]{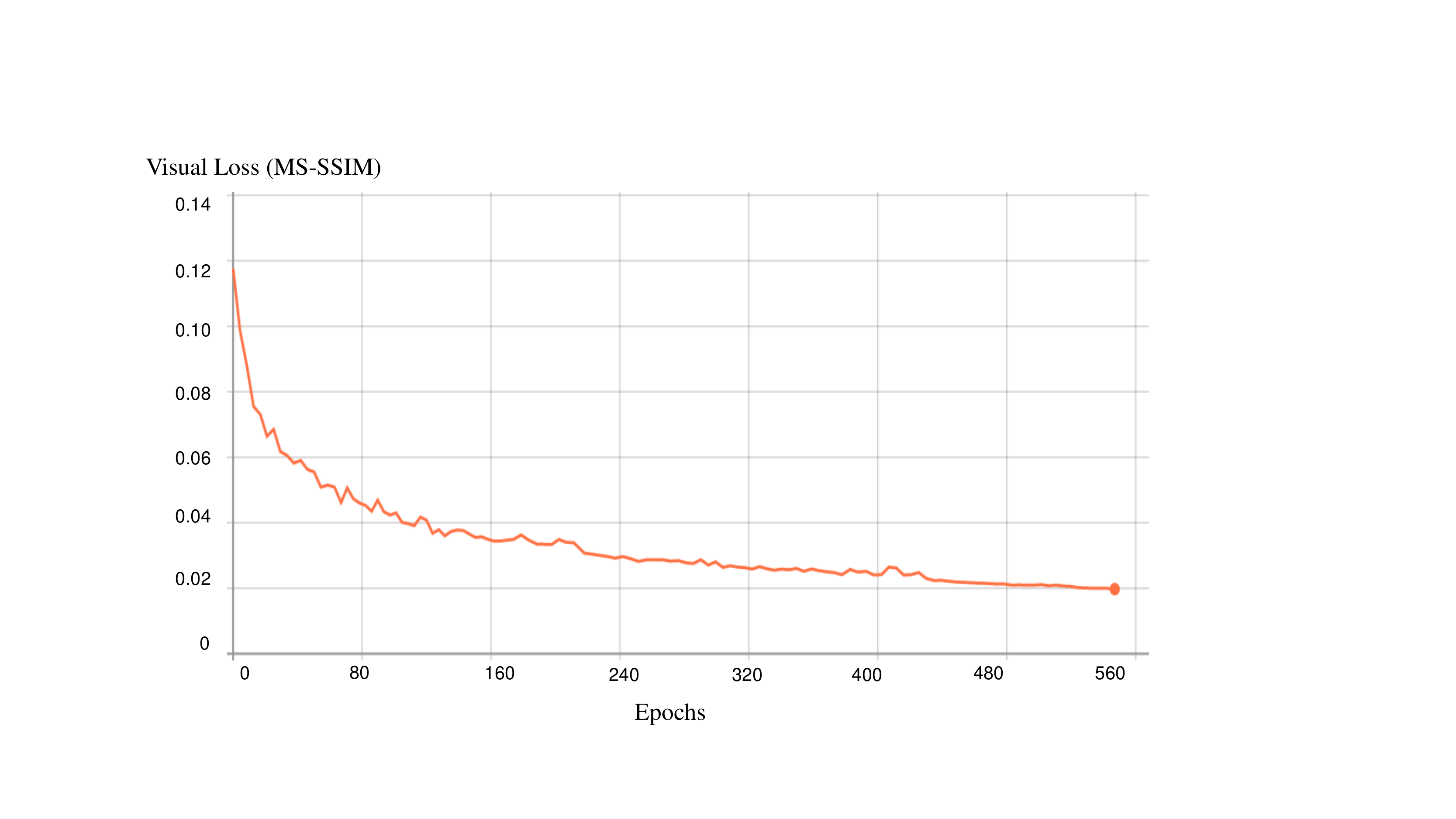}
	\caption{We train the MIAIS framework for large iterations, and show the convergence curve of visual loss (MS-SSIM).}
	\label{converge_a}
\end{figure}

Generally speaking, the identity information consistency of secret image tends to preserve more discriminative features of the secret image, which means the large perturbations within the cover image. Whereas, the visual information consistency of cover image allows small perturbations within the cover image. We measure the identity information consistency with the recognition accuracy (Acc) gap between the container image and the secret image domain, and the visual information consistency with MSE and MS-SSIM. As for the container image domain, the identity information consistency of secret image and visual information consistency of cover image affect and compete each other. In order to evaluate the effectiveness of identity preserving under certain level of visual information hiding, we set a hiding threshold to control the visual quality of steganography. To show the convergence of visual loss (MS-SSIM), we train our baseline for large iterations, without any constraints on identity information consistency or on dynamically adjusting the weights of losses. From Figure~\ref{converge_a}, we observe that the visual loss converges to about 0.02 on the validation dataset, which means that the model is well-performed on visual hiding quality when the visual loss is about 0.02.

\begin{figure}[h]
	\centering
	\subfigure[The influence of hiding threshold on MSE.]{
		\begin{minipage}[t]{0.5\linewidth}
			\centering
			\includegraphics[width=3.0in]{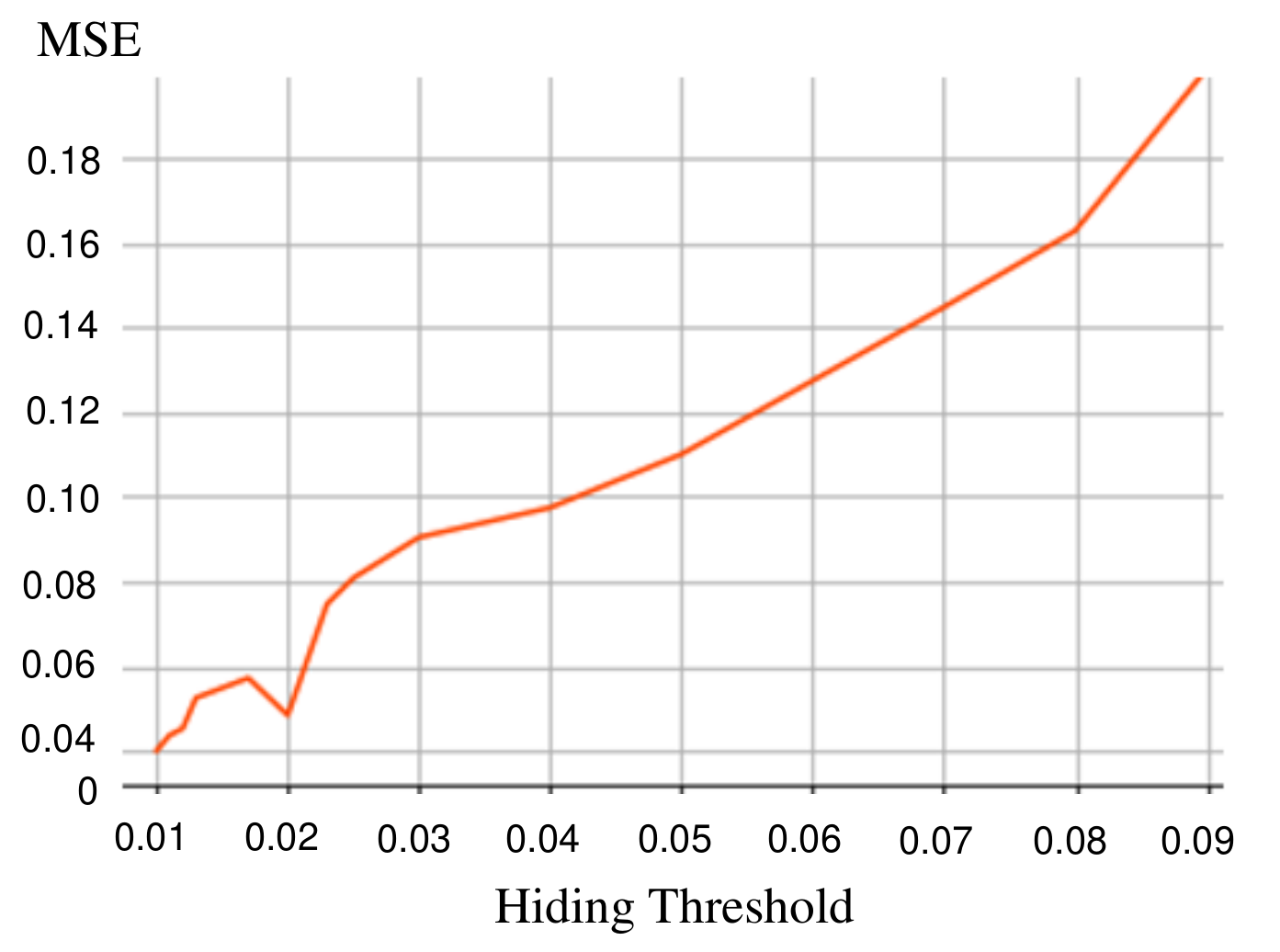}
		\end{minipage}%
	}%
	\subfigure[The container images under different MS-SSIM.]{
		\begin{minipage}[t]{0.5\linewidth}
			\centering
			\includegraphics[width=3.7in]{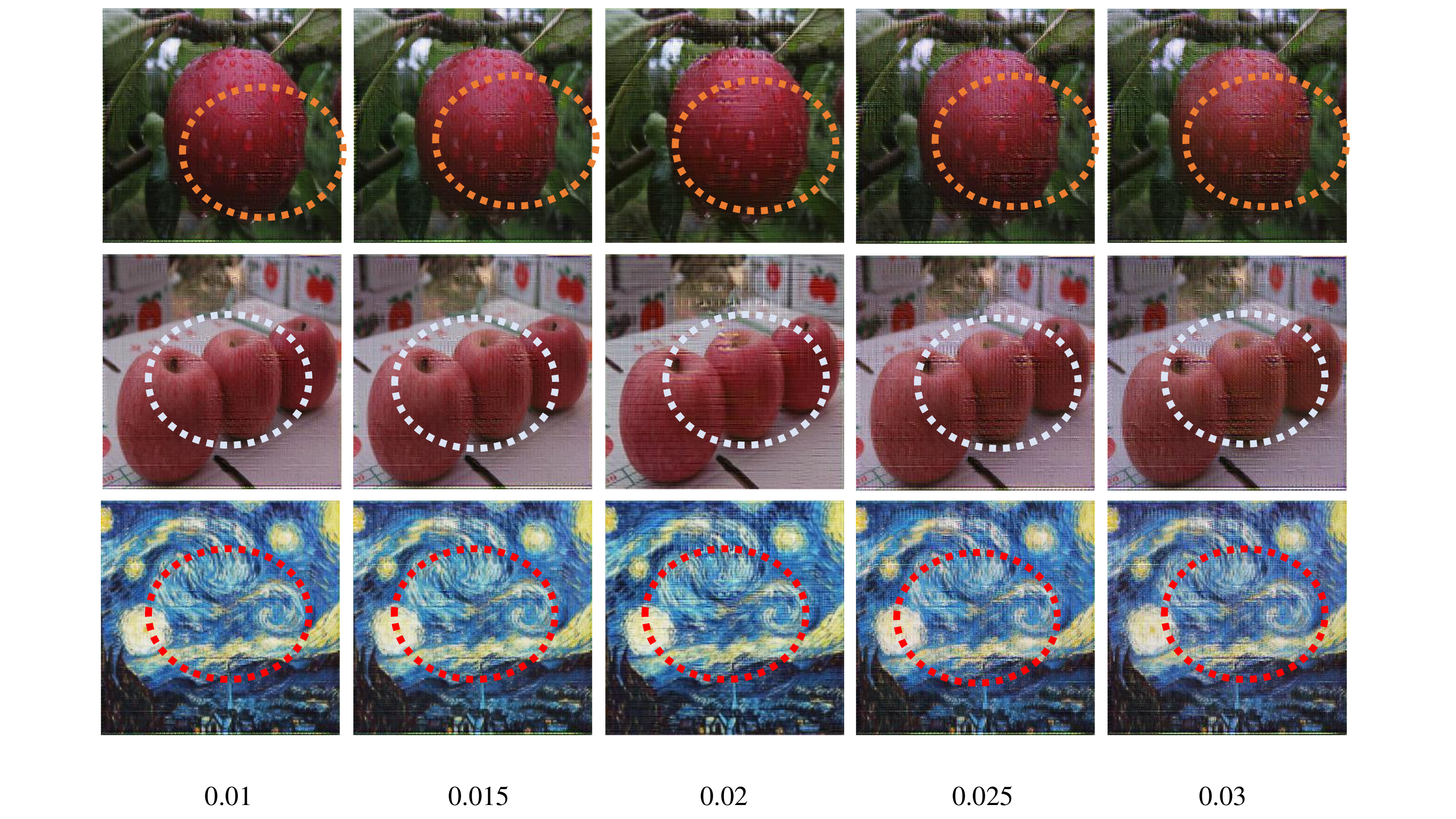}
		\end{minipage}%
	}%
	\caption{(a) We calculate MSE on cover-container image pairs training with different hiding thresholds. (b) Visual quality of container images under different visual loss (MS-SSIM) between the container image and the cover image.}
	\label{converge_b}
\end{figure}

We dynamically adjust the weights of losses in Eqs~(11) and (13) through the magnitude relation between $\mathcal{L}_{vis}$ and the hiding threshold. To show the affect of different hiding thresholds, we train our MIAIS framework with the thresholds around 0.02, and evaluate these models with the MSE between the container image and cover image. As shown in Figure~\ref{converge_b} (a), the threshold of 0.02 is a obvious turning point in the curve of different hiding thresholds. Therefore, in order to ensure a good visual invisible of the generated image, we finally set the hiding threshold to 0.02 in the implementation. As shown in Figure~\ref{converge_b} (b), the overall perturbation of the generated image is not obvious. However, after careful attention to the region in the circles, it can be found that the face slowly emerges with the increase of threshold. Actually, it is ok to set the hiding threshold to values around 0.02, but the higher the hiding threshold, the worse hiding quality will be.

\section{Visual Quality of Different Ablation Experiments} \label{apdx_ablation}

\begin{figure*}[h]
	\centering
	\makebox[\columnwidth][c]{
		\includegraphics[scale=0.75]{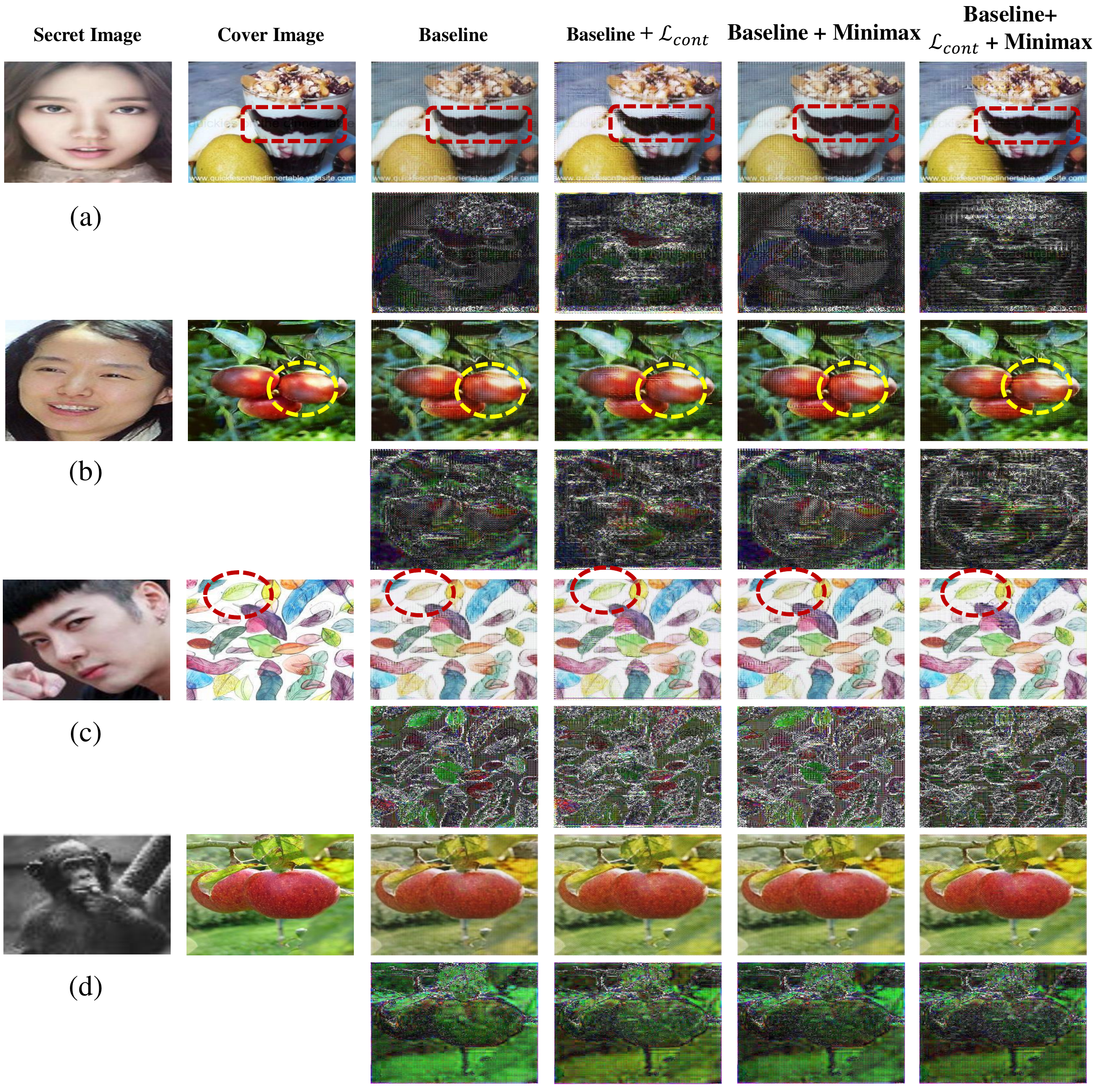}
	}
	\caption{Visual quality of different ablation experiments. The first to third example are from the AFD and the last one is from the Tiny-Imagenet dataset. All the even rows are the residual image between the cover and container image. To amplify the variation in residual image, we enlarge the residual image for 5 times, and use an image processing technique called adaptive histogram equalization on it.}
	\label{ablation_sectionB}
	\vspace{14mm}
\end{figure*}

Figure~\ref{ablation_sectionB} shows the visual quality of different ablation experiments. Through the control of hiding threshold for visual loss, all the generated images have an imperceptible perturbation compared with the original cover image. But it is interesting that, although every method has an almost similar MS-SSIM, their generated images have different styles. The methods with content loss preserve more visual information (tones, textures, and colors) of original cover image. Through the first example (a), we can see the methods without content loss make the generated images contain more tone variation and fewer details highlighted in red dash rectangles. In the second example (b), the textures in yellow dash circles turn to more smooth and regular with the content loss. In the third example (c), the color of green leaves highlighted in red dash circles is preserved by the methods with content loss. As shown in residual image of the first and second examples, we find that the methods without content loss expose more face sketch information from the original secret image. From cover images in Figure~\ref{ablation_sectionB}, it is worth noticing that the cover images that have more complex textures could get better visual hiding performance (the third example (c)).

\begin{figure}
	\centering
	\makebox[\textwidth][c]{\includegraphics[scale=0.35]{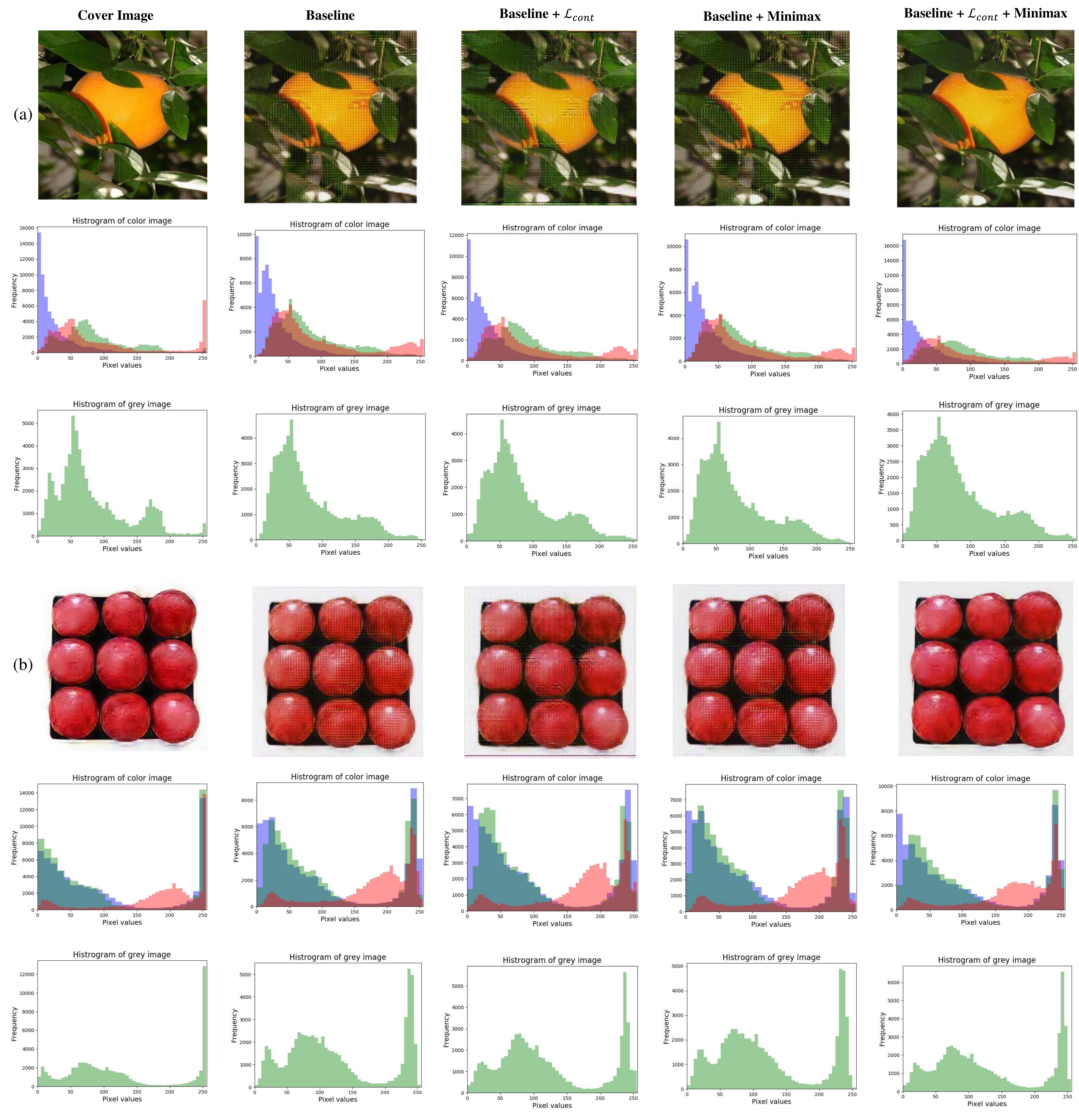}}
	\caption{Visual quality comparison between different ablation experiments. The first to third rows of each examples are original images, histograms of color images, histograms of grey images separately.} 
	\label{ablation_visual_information_c}
\end{figure}

Moreover, we adopt the color histograms (RGB) of images to visualize the color distribution among the ablative methods, as shown in \mbox{\cref{ablation_visual_information_c}}. From the color histograms of example (a) and (c) in \mbox{\cref{ablation_visual_information_c}}, the content loss \textbf{$\mathcal{L}_{cont}$ + Minimax} could make the color distribution of the container image more consistent with the cover image. From the grey histograms in \mbox{\cref{ablation_visual_information_c}}, the content loss \textbf{$\mathcal{L}_{cont}$} could smooth the color distribution of the container image, especially from the example (b). 

\clearpage
\section{Visual Quality of Visual Information Hiding Methods} \label{apdx_VQ_others}

\begin{figure}[h]
	\centering
	\includegraphics[width=.75\textheight]{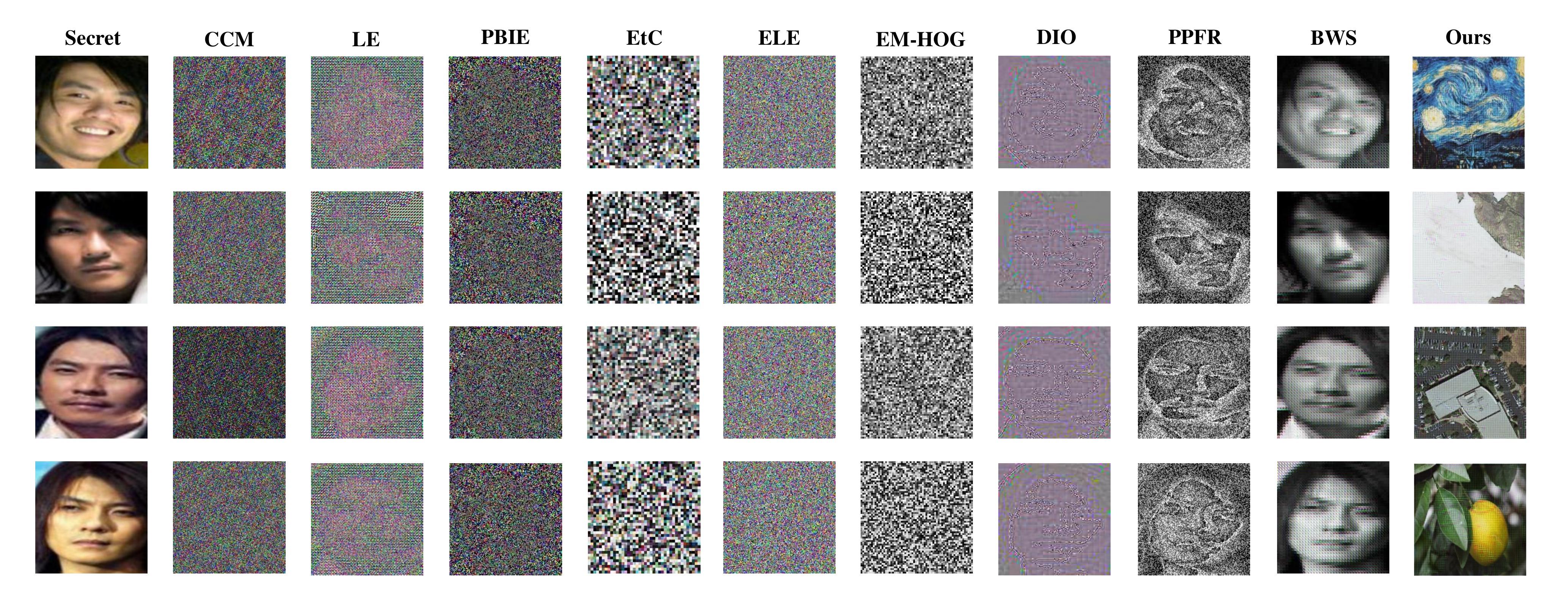}
	\caption{Comparison with visual information hiding methods in visual quality. The container images produced by our MIAIS are more \textbf{realistic} than the other methods. BWS only naively shuffles the 4 * 4 pixel blocks without consideration for the privacy presercing.}
	\label{comparison-other}
\end{figure}

In this section, we supplement the visual quality of other visual information hiding methods compared with the proposed \textbf{MIAIS} framework, as shown in Figure~\ref{comparison-other}. \textbf{CCM}~[23] is a block image encryption method based on hybrid chaotic maps, and generated key streams are dependent on each secret image. So we can see that the encrypted image is chaotic and random, and the details and features of the secret picture are invisible. 
\textbf{LE}~[27], \textbf{EtC}~[24], \textbf{ELE}~[13] and \textbf{EM-HOG}~[26] are all based on block-wise image scrambling, and their block size is 4*4. They adopt some random operations, including block rotation \& inversion, negative-positive transform, color component shuffling, etc. Among them, the key of \textbf{LE} is simple, so we can see that few outline of the face in the encrypted image. The other three encrypted images are so messy that we can't see useful information at all. 
\textbf{PBIE}~[29] is a pixel-based image encryption method, which adopts negative-positive transform. The visual information of the secret image is destroyed in pixel-level. 
\textbf{DIO}~[30] adopts the P3 algorithm with a threshold of 20, it works by removing the DC and large AC coefficients from the public version of image. Obviously, most of the sensitive information is removed, but it is also possible to see the outline of the face. 
In \textbf{PPFR}~[12], every pixel of the input image is gone through a bitwise XOR operation. We can not only see the outline of the face, but also some details of the facial features in the encrypted image. 
\textbf{BWS} only naively shuffles the 4 * 4 pixel blocks so that can't meet the encryption requirements at all. 
The encrypted image obtained by the proposed \textbf{MIAIS} framework is visually the same as the cover image and completely different from the secret image, which can achieve a great encryption effect.

\section{Experiments with High PSNR}
The scope of our work is to achieve a trade-off between direct recognition measured by the accuracy and visual concealment measured by PSNR. In the experiments, recognition accuracy was our first priority, which possibly sacrificed the PSNR performance to a certain extent. Actually, the PSNR result is controlled by the hiding threshold (default T = 0.02). To raise the PSNR result, we adjust the hiding threshold to 0.001, and evaluate the performance under the direct recognition and multi-task scenario, as shown in \mbox{\cref{Table_cover_robustness_high_PSNR}} and \mbox{\cref{Table_Comparison_Steganography_high_PSNR}} separately. 

\input{../tables/robust_cover_high_PSNR.tex}

\input{../tables/comp_steg_other_high_PSNR.tex}

From \mbox{\cref{Table_cover_robustness_high_PSNR}}, when the PSNR results are around 30db, the recognition accuracy is reduced by 1.6\% on average compared with our original recognition version, which means that our method is still capable of hiding the discriminative information. From \mbox{\cref{Table_Comparison_Steganography_high_PSNR}}, although we can force the PSNR results of other methods around 30db, the recognition accuracy of HliPS and HiDDeN are lower than our method with a large margin, and some methods (SteganoGAN, StegNet) even fail to perform recognition.

\section{Experiments on the Tiny-AFD Dataset} \label{tiny_set}

To explore the real-world biometric recognition scenario, we reduce the number of training images by sampling 10 faces per identity. We evaluate the performance of our method under the multi-task scenario. As shown in \mbox{\cref{tinyset}}, Our method also achieves good performance in the real-world biometric recognition scenario with around 10 images per identity. Given the similar PSNR, our method achieves the best recognition accuracy (the odd line). Given the similar recognition accuracy, the PSNR of our method is higher than the SteganoGAN and StegNet with a large margin.

\input{../tables/tiny_dataset_with_restorer.tex}

\section{Modification in an Image}

\begin{figure}[h]
	\vspace{10mm}
	\centering
	\makebox[\textwidth][c]{\includegraphics[scale=0.66]{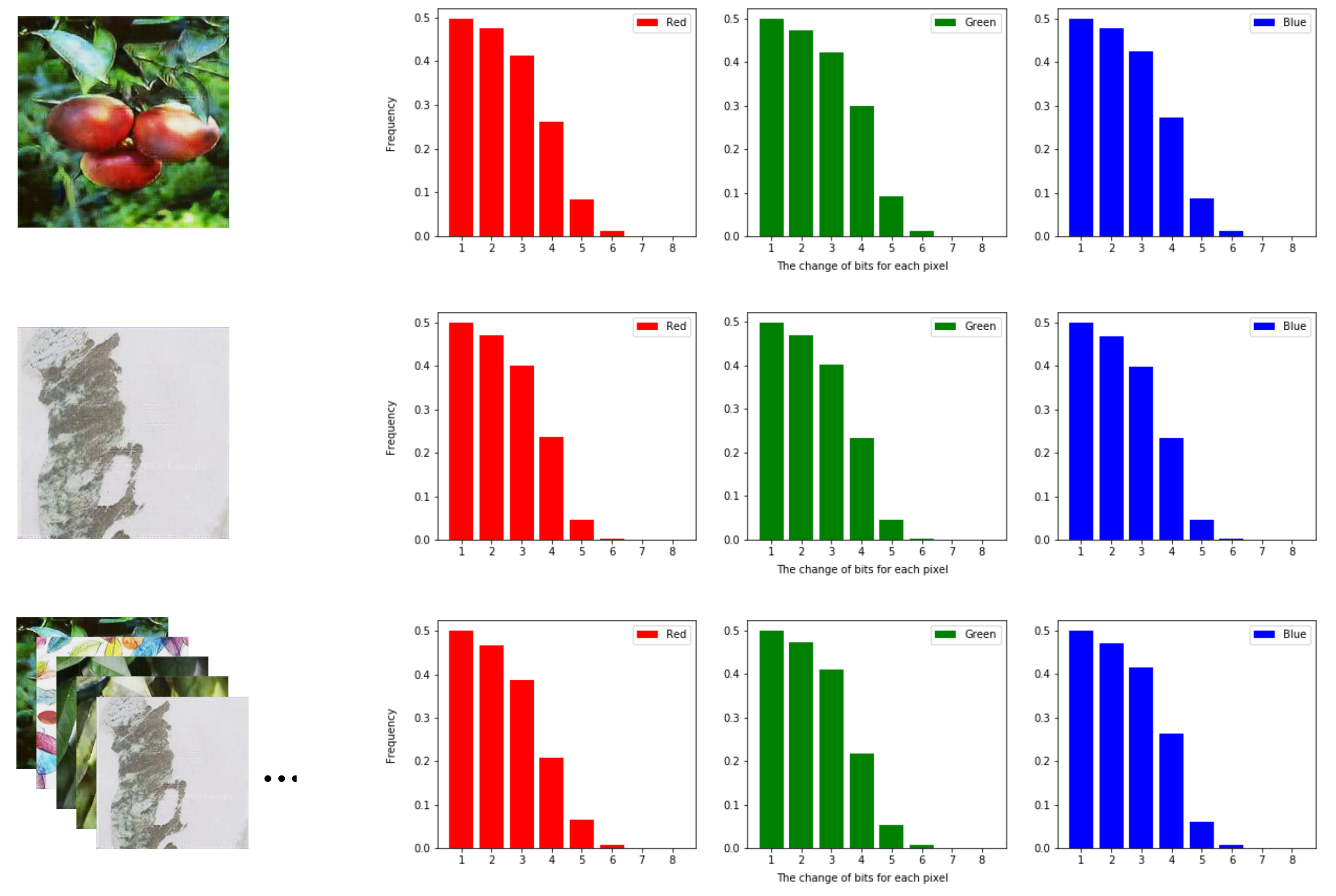}}
	\caption{The frequency histograms of bit perturbation on container images on AFD dataset. The left part shows some examples. The right part shows the corresponding bit perturbation between the container image and the corresponding cover image at each 24 bits (8 $\times$ (R, G, B)) of each pixel. We draw the frequency histograms to visualize the bit perturbation. Note that ``8'' indicates the most significant bit, and ``1'' indicates the least significant bit (LSB). The framework changes not only the LSB parts, but also some subsequent significant bits. The frequency of bit perturbation increases as the bit significance decreases, because the perturbation of the significant bit indicates a large variation in visual appearance of the image.} 
	\label{bit_perturbation}
\end{figure}

\begin{figure*}[h]
	\centering
	\makebox[\textwidth][c]{\includegraphics[scale=0.96]{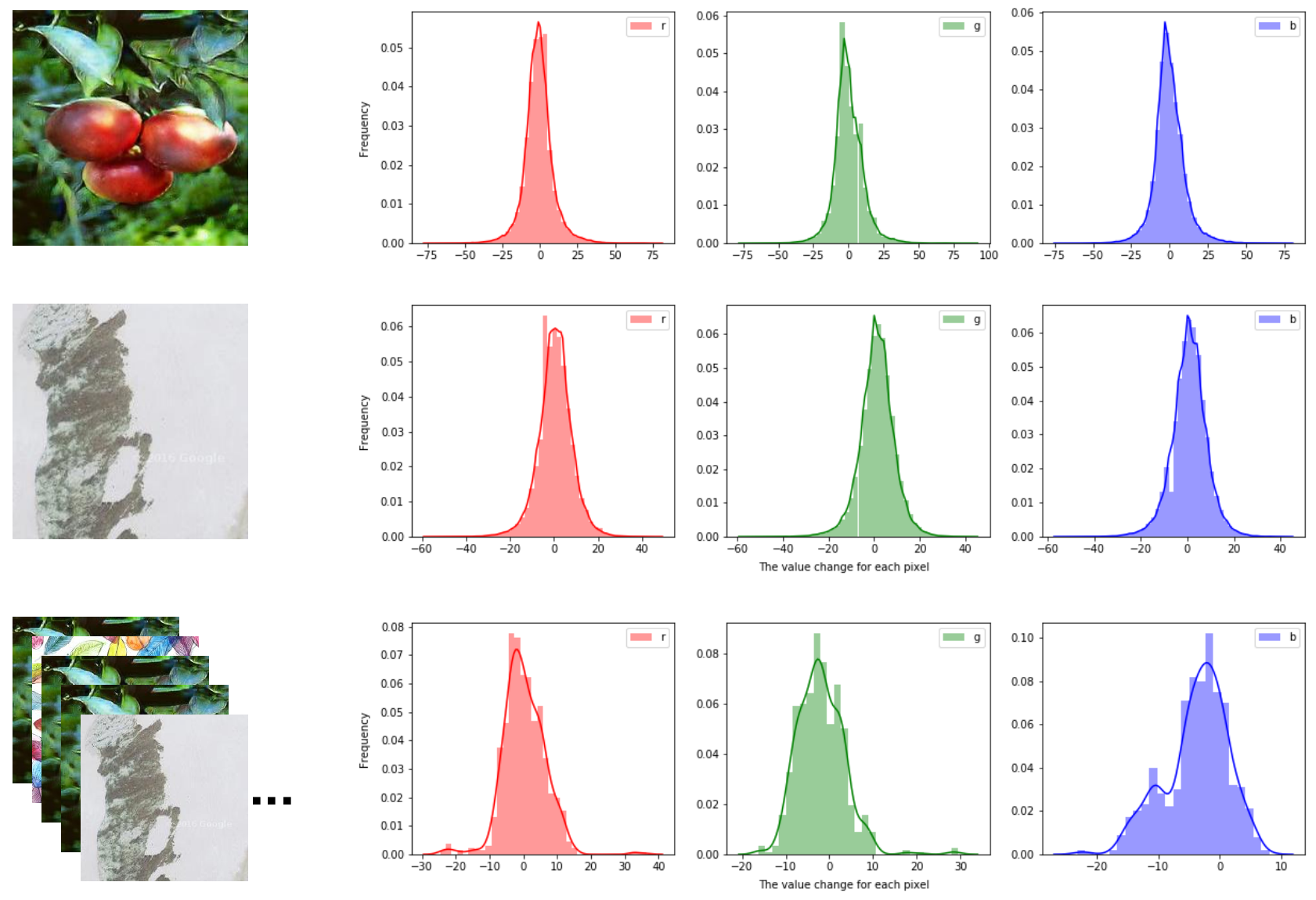}}
	\caption{The frequency histograms of pixel value modifications on container images on AFD dataset. For some examples and 1000 test examples, we count the number of pixel value modifications between the container image and the corresponding cover image in each RGB channel. We draw the frequency histograms to visualize the modifications. From the average frequency histogram of 1000 test examples, a large amount of the pixel value modifications are between -10 and 10, which means that the framework conceals the secret image through small perturbation (10 pixel value) on the cover image. } 
	\label{pixel_modification}
\end{figure*}

Our framework directly generates the container image on the whole by the deep learning-based steg-generator instead of modifying the least significant bit (LSB). To further show the modification of our steganography network, we do statistics on the distribution of perturbation for each bit and the distribution of the modification values of pixels, as shown in \mbox{\cref{bit_perturbation}} and \mbox{\cref{pixel_modification}}, respectively.

1) \textbf{\textit{Bit perturbation:}} Obviously, our method modified among the different bit channel parts instead of only the LSB parts. Moreover, the LSB method is incapable of hiding a secret image into another full-size cover image because of the low payload. From \mbox{\cref{bit_perturbation}}, the framework changes not only the LSB parts, but also some subsequent significant bits. The frequency of bit perturbation increases as the bit significance decreases, because the perturbation of the significant bit indicates a large variation in visual appearance of the image.

2) \textbf{\textit{Pixel modification:}} We calculate pixel value modifications between the container image and the corresponding cover image. As shown in \mbox{\cref{pixel_modification}}, the distributions of pixel value modifications in the three channels are approximately Gaussian, which means that the framework hides the information in an image through perturbing a small amount of pixel values (about 10 on average).

\clearpage
\section{Visual Quality of Different Image Steganography Methods} \label{apdx_VQ_MTL}

\begin{figure}[h]
	\centering
	\includegraphics[width=.75\textheight]{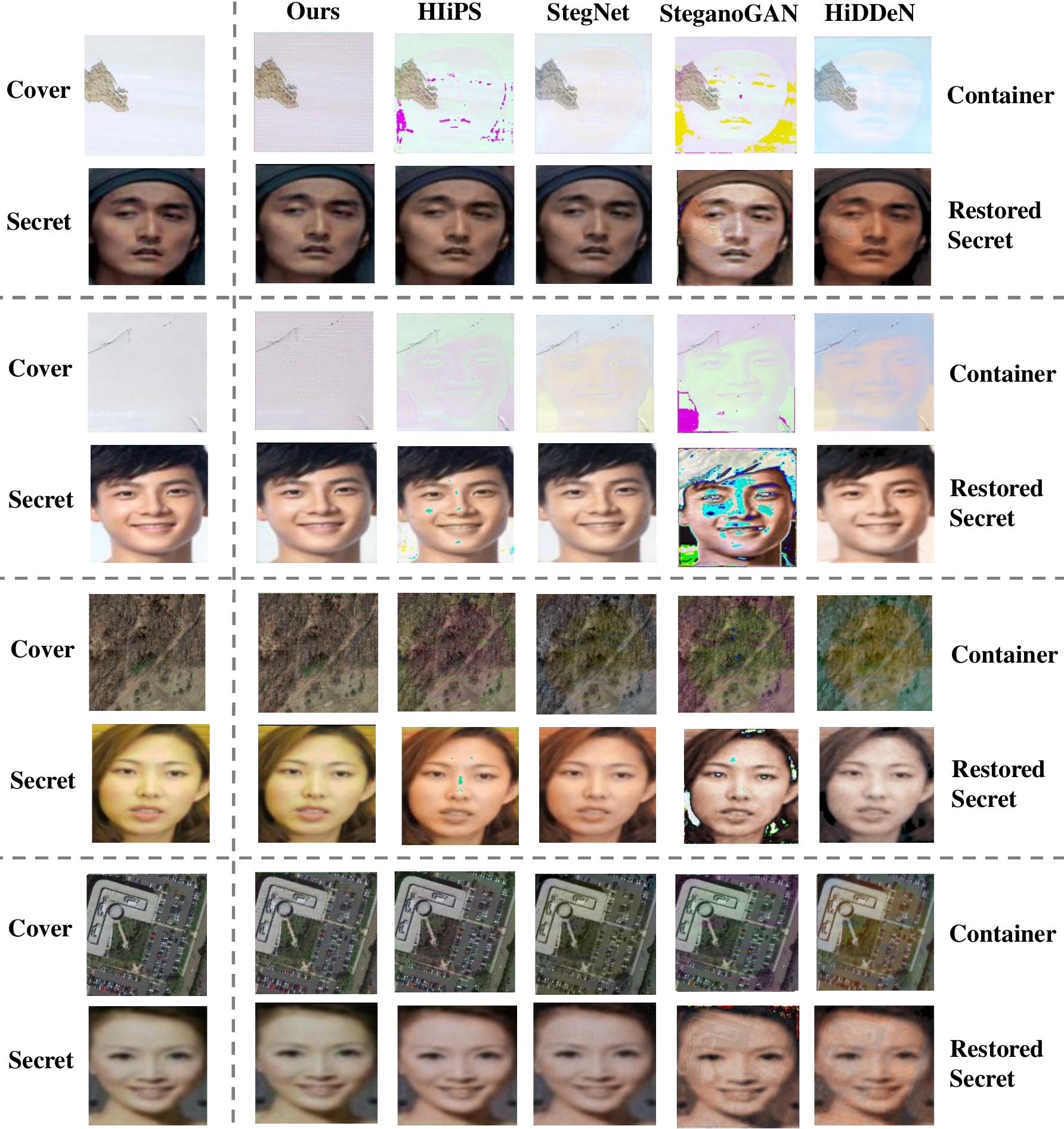}
	\caption{Comparison with different image steganography methods.}
	\label{comparison-other_1}
\end{figure}

\begin{figure}[h]
	\centering
	\includegraphics[width=.75\textheight]{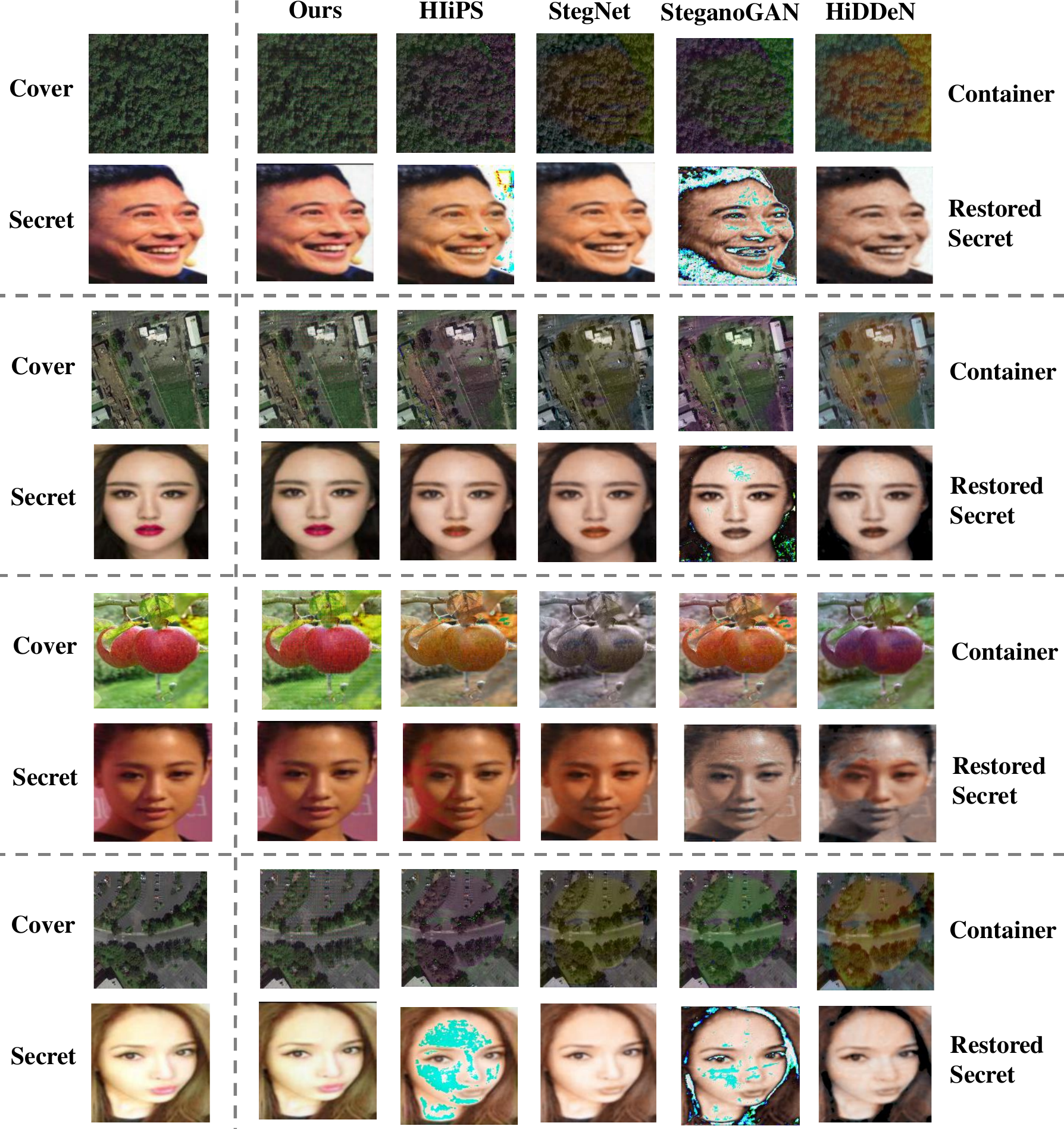}
	\caption{Comparison with different image steganography methods.}
	\label{comparison-other_2}
\end{figure}

\begin{figure}[h]
	\centering
	\includegraphics[width=.75\textheight]{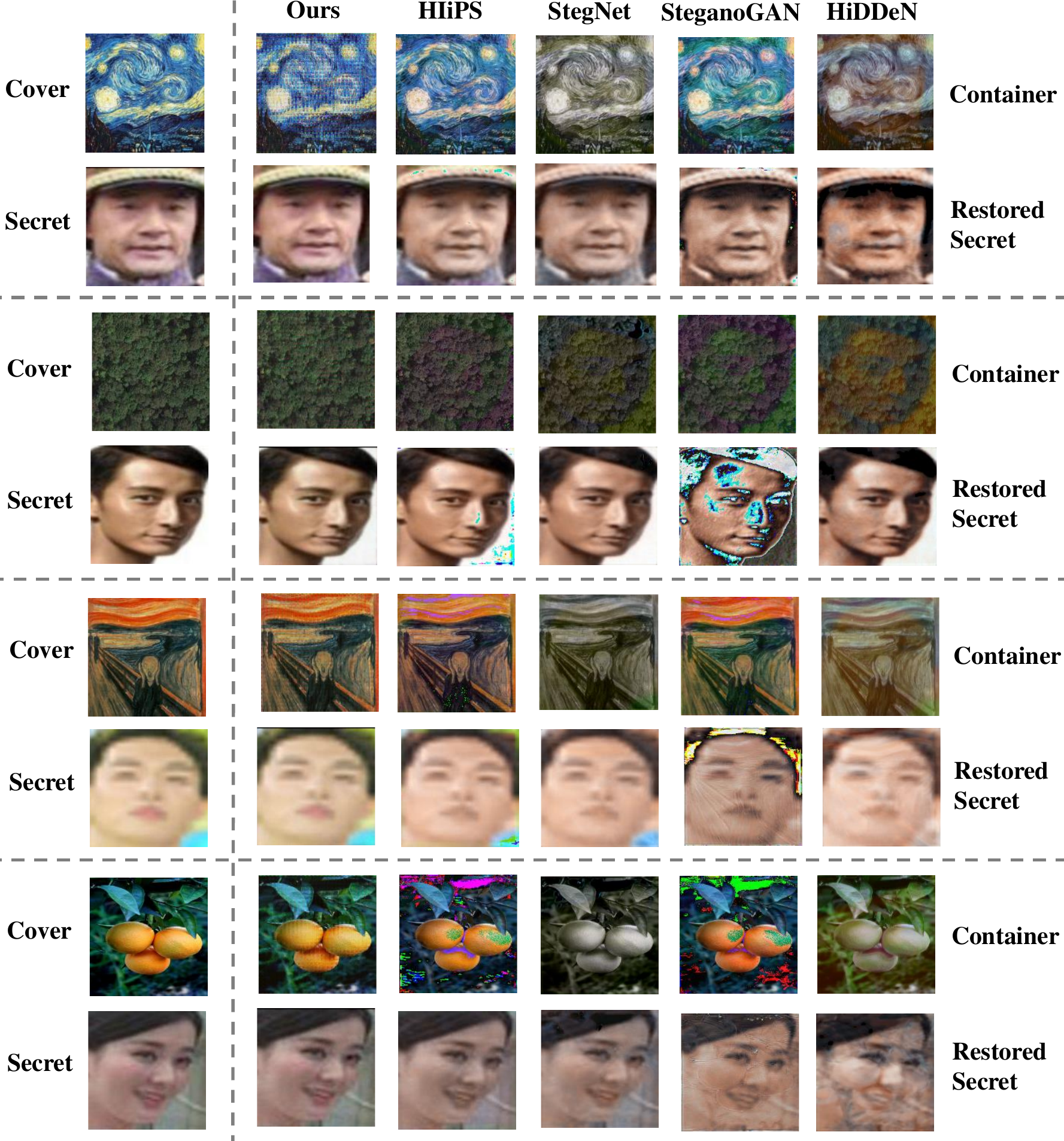}
	\caption{Comparison with different image steganography methods.}
	\label{comparison-other_3}
\end{figure}

In this section, we supplement the visual quality of different image steganography methods compared with the proposed \textbf{MIAIS} framework, as shown in \cref{comparison-other_1,comparison-other_2,comparison-other_3}. Compared with our MIAIS, the container images generated by the other methods have a darker tone (HiDDeN, StegNet) for the paintings and some color noise pixels (SteganoGAN, HIiPS) for real photo images. From cover images in \cref{comparison-other_1}, it is worth noticing that the cover images that have more complex textures could get better visual information hiding performance.

\newpage

\newpage

\clearpage

\section{Network Architecture Details}

\begin{figure}[h]
	\centering
	\makebox[\textwidth][c]{\includegraphics[scale=0.85]{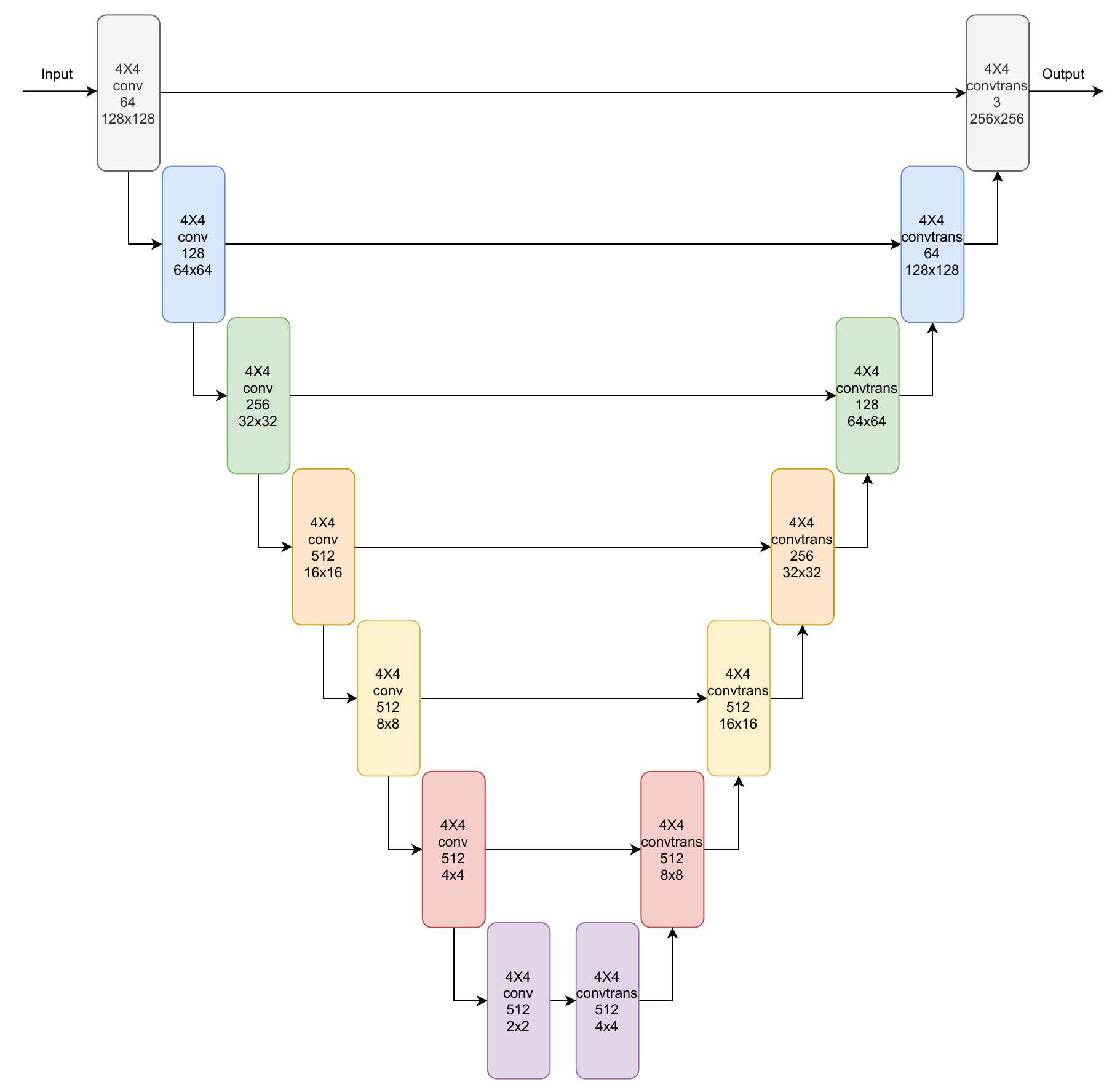}}
	\caption{The network architecture of the steg-generator.} 
	\label{steg_generator}
\end{figure}

\begin{figure}[h]
	\centering
	\makebox[\textwidth][c]{\includegraphics[scale=0.85]{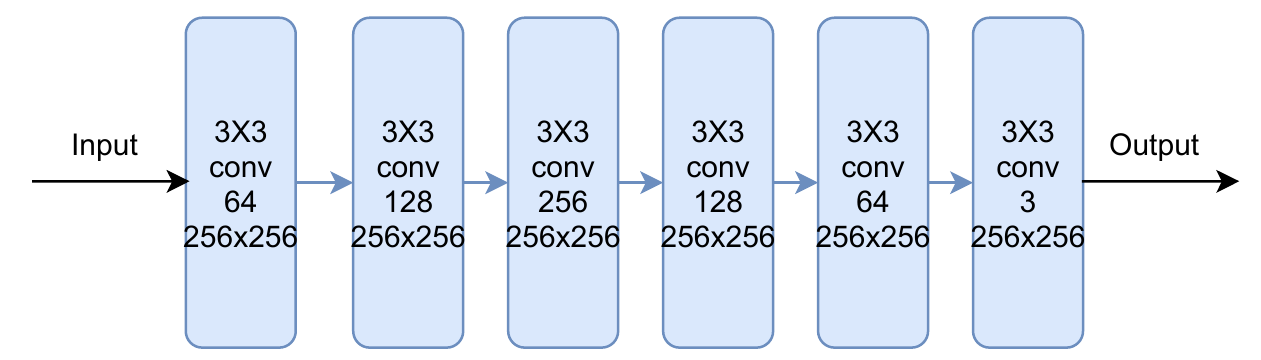}}
	\caption{The network architecture of the steg-restorer.} 
	\label{steg_restor}
\end{figure}

\begin{figure}[h]
	\centering
	\makebox[\textwidth][c]{\includegraphics[scale=0.79]{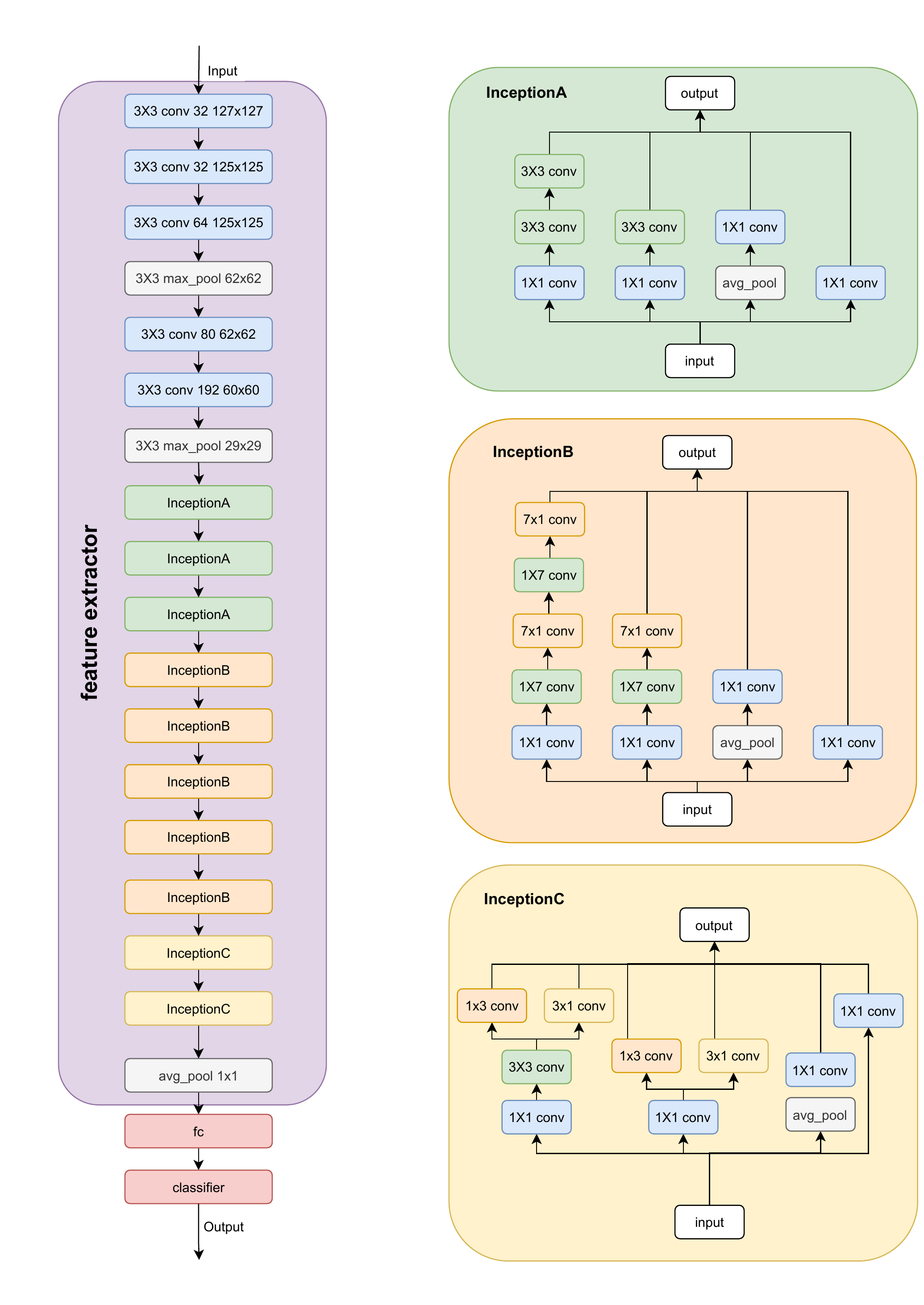}}
	\caption{The network architecture of the steg-classifier.} 
	\label{steg_classifier}
\end{figure}

As described in the Section~IV-A2 of the main text, we draw the network architecture details (\mbox{\cref{steg_classifier,steg_generator,steg_restor}}).

\ifCLASSOPTIONcaptionsoff
  \newpage
\fi

%% file: sections/abstract.tex
\begin{abstract}

High-capacity image steganography, aimed at concealing a secret image in a cover image, is a technique to preserve sensitive data, e.g., faces and fingerprints. Previous methods focus on the security during transmission and subsequently run a risk of privacy leakage after the restoration of secret images at the receiving end. To address this issue, we propose a framework, called Multitask Identity-Aware Image Steganography (MIAIS), to achieve direct recognition on container images without restoring secret images. The key issue of the direct recognition is to preserve identity information of secret images into container images and make container images look similar to cover images at the same time. Thus, we introduce a simple content loss to preserve the identity information, and design a minimax optimization to deal with the contradictory aspects. We demonstrate that the robustness results can be transferred across different cover datasets. In order to be flexible for the secret image restoration in some cases, we incorporate an optional restoration network into our method, providing a multitask framework. 
The experiments under the multitask scenario show the effectiveness of our framework compared with other visual information hiding methods and state-of-the-art high-capacity image steganography methods.

\end{abstract}

%% file: sections/intro.tex
\section{Introduction}
\begin{figure}[t]
	\centering
	\includegraphics[width=.36\textheight]{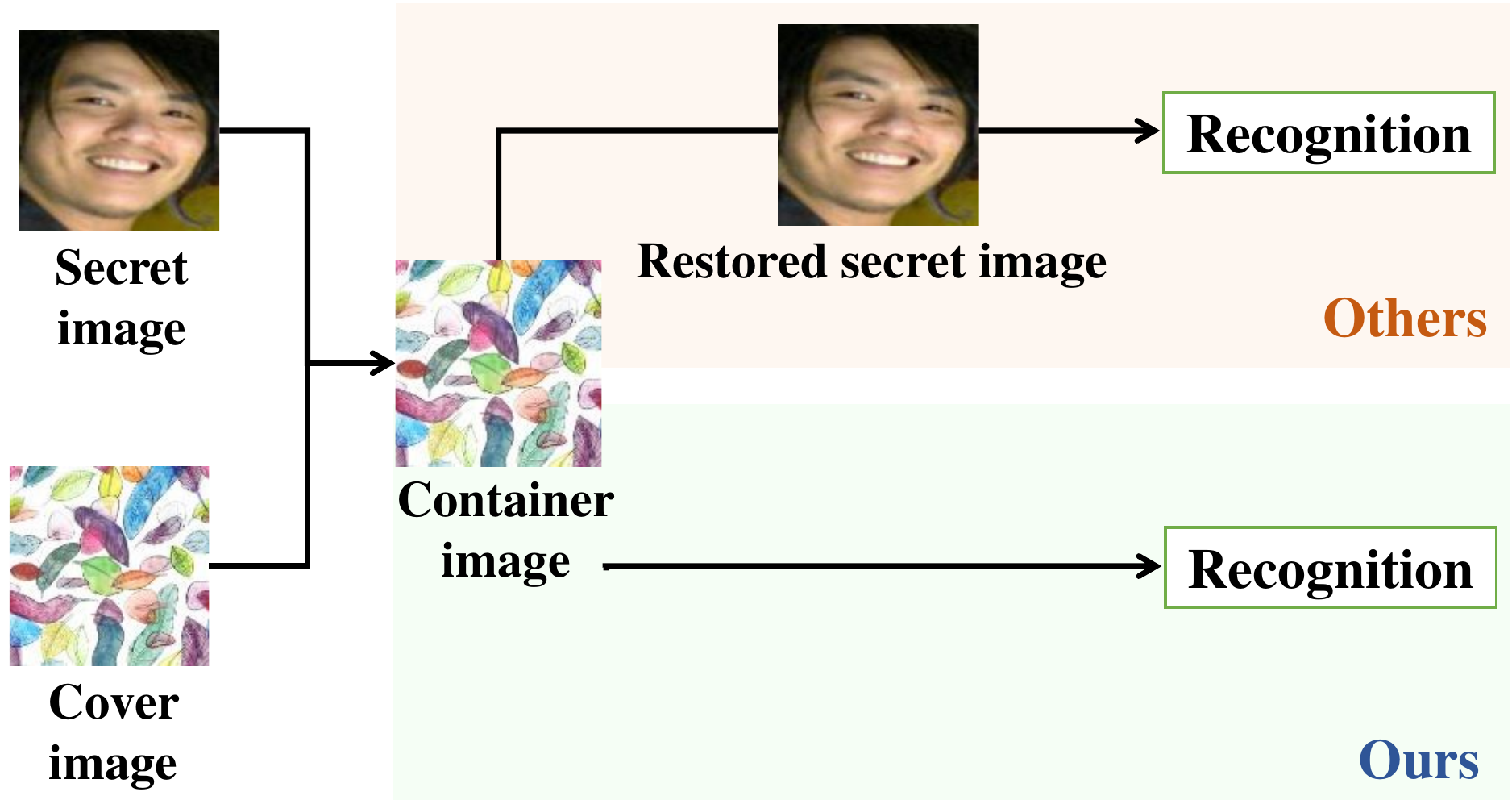}
	\caption{Comparison of previous image steganography methods and ours. An image steganography method generates a container image by hiding a secret image in a cover image. The container image would be transferred from a sender to a receiver. Our difference from previous methods lies in the processing of container images. \textbf{Top}: Previous methods restore secret images from container images for recognition, which raises a risk of privacy leakage. \textbf{Bottom}: Our framework performs recognition directly on container images.}
	\label{intro-difference}
\end{figure}

\IEEEPARstart{V}{isual} security authentication, e.g., face recognition~\cite{galbally2013image,huang2015benchmark,xie2017robust} and fingerprint identification~\cite{valdes2019review}, has achieved considerable advances in recent years. Its wide application also poses a challenge that such sensitive data as face and fingerprint need to be protected. 
High-capacity image steganography, which generates a container image to conceal a secret image in a cover image, is an elegant and widespread technique to address this issue~\cite{hussain2018image}. 
Previous methods~\cite{wu2015steganography,baluja2017hiding} focus on the protection of secret images during transmission. 
Consequently, they run a risk of visual privacy leakage after restoration: once the receiver is under attack, secret images can be stolen.
To this end, we propose a framework, called Multitask Identity-Aware Image Steganography (MIAIS), to perform recognition directly on container images without restoring secret images, as shown in \cref{intro-difference}.

To perform direct recognition on container images, two intuitively contradictory issues need to be addressed. 
On the one hand, concealment allows \emph{small} perturbations in cover images, which means the container images should look similar to their cover images. 
On the other hand, direct recognition may require \emph{large} perturbations in cover images to preserve discriminative features of secret images. 
The former aspect is the focus of prevalent image steganography; we focus on the latter aspect and propose a strategy to deal with the contradiction. 


\begin{figure*}
	\centering
	\makebox[\textwidth][c]{\includegraphics[scale=0.56]{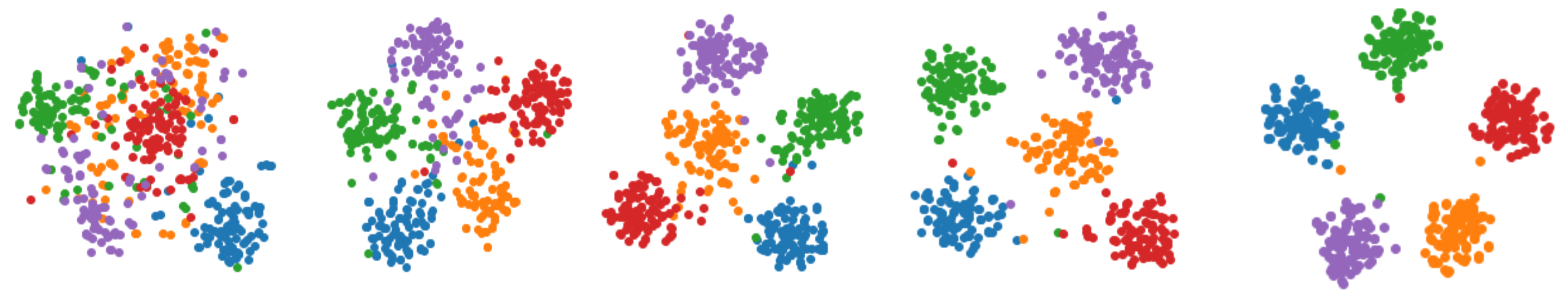}}
	\leftline{\footnotesize\hspace{1.1cm}(a) HIiPS~\cite{baluja2017hiding} \hspace{1.4cm}(b) SteganoGAN~\cite{zhang2019steganogan} \hspace{0.5cm}(c) With the content loss \hspace{0.4cm}(d) With the minimax optimization \hspace{1.2cm} (e) With (c) and (d)}
	\caption{T-SNE~\cite{dermaaten2008visualizing} visualization of different methods in terms of high-level features of container images out of five classes on the AFD dataset~\cite{xiong2018an}. (a) HIiPS~\cite{baluja2017hiding} and (b) SteganoGAN~\cite{zhang2019steganogan} are two mainstream image steganography methods, where the features are not discriminative. (c) Our content loss preserves the identity information to produce discriminative features on container images. (d) The minimax optimization can also produce discriminative features on container images. (e) When both the content loss and the minimax optimization are used, the features are even more discriminative. Best viewed in color.} 
	\label{intro_tsne}
\end{figure*}

For direct recognition, we introduce a content loss, adding a similarity constraint between a secret image and its container image. 
In general, identity information lies in high-level features in a deep network~\cite{he2016deep}. 
Thus, we adopt the feature extractor part of a classifier to implement a similarity constraint for high-level features. 
The content loss preserves the discriminative high-level feature consistency between the container and secret image, although the container is similar to the cover image visually.
As a comparison, previous steganography methods, such as HIiPS~\cite{baluja2017hiding} and SteganoGAN~\cite{zhang2019steganogan}, are not suitable for direct recognition because of their indiscriminate features, as shown in \cref{intro_tsne}~(a), (b), and (c).  


To deal with the aforementioned contradiction, 
we design a minimax optimization including a network for hiding secret images called a \textbf{steg-generator} and two container image classifiers called \textbf{steg-classifiers}, as shown in \cref{pipeline}. 
The steg-generator and the steg-classifiers are alternately trained to compete with each other. 
Specifically, at Stage A, the two steg-classifiers are trained to maximize the discrepancy of them with the steg-generator fixed. At Stage B, with the two steg-classifiers fixed, the steg-generator is trained to minimize the discrepancy of the two steg-classifiers.
Stage A and Stage B are performed iteratively.
As a result, the process decreases the intra-class distance and increases the inter-class distance, obtaining more discriminative container images for recognition, as shown in \cref{intro_tsne}~(d) and (e).

Although our method can perform recognition without restoration, in order to be flexible for the secret image restoration in some cases, we incorporate an optional restoration network called a \textbf{steg-restorer} into our pipeline, providing a multitask learning scenario in our framework, as shown in \mbox{\cref{pipeline}}. 
It is extremely easy to implement it by adding a visual similarity constraint between secret images and the restored secret images in container images. So far, our MIAIS framework could entirely generate a container image one path through an end-to-end learning pipeline, while maintaining a balance among the \textbf{visual concealment}, \textbf{identity recognition}, and \textbf{secret restoration} tasks.


The main contributions are summarized as follows:
\begin{itemize}
	\item We propose a novel framework to perform direct recognition on container images, preventing privacy leakage while simplifying the recognition workflow. To the best of our knowledge, it is the first work to propose and study the direct recognition task on container images.
	\item We introduce a content loss to preserve identity information for secret images. 
	\item We design a minimax optimization handling the intuitive contradiction between preserving identity information and making container images similar to the corresponding cover images.
	\item We conduct the extensive experiments to show the 1) effectiveness of our framework compared with other visual information hiding methods, 2) robustness of the recognition results across different cover sets, and 3) better performance compared to the state-of-the-art high-capacity image steganography methods.
\end{itemize}

%% file: sections/related.tex
\section{Related Work}

In this section, we first review traditional visual information hiding methods used for privacy protection. Then, we focus on image steganography. Finally, we review some adversarial learning strategies, including generative adversarial networks and minimax optimization.

\subsection{Visual Information Hiding}
\label{Visual_Information_Hiding}

With the introduction of recognition, many works have been considering the security computation on sensitive images or features~\cite{ergun2014privacy,madono2020block,hu2016securing}, in which visual information is hidden from humans in the whole process. 

A group of the visual information hiding works focused on \textit{cryptography}~\cite{gilad2016cryptonets, hesamifard2017cryptodl, xu2019cryptonn}. In order to protect the data at the receiving end, homomorphic encryption (HE) was introduced as the underlying mechanism. Several privacy-preserving approaches~\cite{gilad2016cryptonets,yonetani2017privacy, wang2018efficient,sadeghi2009efficient,gentry2009fully} hid the visual information in encrypted data and applied the neural networks to recognize the encrypted data. However, it was not feasible due to the computational complexity and memory costs~\cite{madono2020block}. Furthermore, the mathematical operations (additive and multiplicative) in HE are limited, which cannot be adapted to complex transformations in state-of-the-art deep neural networks.

Now that the cryptography-based methods were not practicable, another group of the research turned to the combined approaches of \textit{cryptography, machine learning}, and \textit{image processing}. 
Ergun~\textit{et al.}~\cite{ergun2014privacy} proposed a Privacy Preserved Face Recognition (PPFR) framework, which encrypted the plain-text by random cryptographic keys based on the continuous chaotic system~\cite{ergun2011high} and directly recognized the encrypted data. Block-based encryption methods were popular in this area, such as Combined Cat Map (CCM)~\cite{wang2015novel} based on hybrid chaotic maps and dynamic random growth technique, encryption method of histograms of oriented gradients (HOG) features~\cite{kitayama2019hog}, and Encryption-then-Compression (EtC) framework~\cite{kawamura2020privacy}. Moreover, Chuman~\textit{et al.}~\cite{chuman2018encryption} presented a block scrambling-based encryption scheme to enhance the security of EtC framework with JPEG compression. 


Several studies had also used \textit{Deep Neural Networks (DNNs)}. Tanaka~\textit{et al.}~\cite{tanaka2018learnable} applied learnable encryption (LE) images to DNNs by reducing the influence of image encryption by adding an adaptation network prior. 
Extended learnable encryption (ELE)~\cite{madono2020block} hid perceptual information by Block-wise image scrambling. 
Sirichotedumrong~\textit{et al.}~\cite{sirichotedumrong2019privacy} extended their conference version~\cite{sirichotedumrong2019pixel} and proposed a pixel-based image encryption method to maintain the important features of original images for privacy-preserving DNNs' classification. 
McPherson~\textit{et al.}~\cite{mcpherson2016defeating} empirically showed how to train artificial neural networks to successfully identify faces and recognize handwritten digits even if the images were protected by various forms of obfuscation techniques, such as mosaicing, blurring~\cite{hill2016effectiveness}, and Privacy Preserving Photo Sharing (P3)~\cite{ra2013p3}. 

Our approach is based on the high-capacity image steganography to generate more realistic images for transmission. 
We compare our method with these methods in \cref{Experiments_Recognition}.

\subsection{Image Steganography}
\label{Related_Work_Steganography}

Steganography is the art and science of hiding secret information in a payload carrier (cover) and get a container~\cite{4655281}. 
It can be categorized by the cover forms to image~\cite{hussain2018image}, audio~\cite{djebbar2012comparative}, video~\cite{sadek2015video}, text~\cite{liu2015text}, DNA~\cite{santoso2015information}, etc. The most popular medium is the image because it has the biggest amount in our life. In this paper, we focus on the high-capacity image steganography where the secret information is also images.

\subsubsection{Traditional Image Steganography}
The secret information of traditional image steganography is usually messages (e.g., text, binary string). From the perspective of embedding domain, traditional steganography algorithms can be divided into spatial-domain algorithms\cite{lie1999data,pevny2010using,pevny2010steganalysis,holub2014universal,holub2012designing,tamimi2013hiding} and transform-domain algorithms~\cite{zhang2009high,ramkumar1999robust,quan2009high}.
The spatial-domain algorithms embed the secret message through pixel brightness value, color, texture, edge, and contour modification of the cover image. The transform-domain algorithms embed secret messages in the transform domain of the cover image under different kinds of transforms, such as Discrete Wavelet Transform (DWT)~\cite{zhang2009high}, Discrete Fourier Transform (DFT)~\cite{ramkumar1999robust}, Discrete Cosine Transform (DCT)~\cite{quan2009high} and so on.

\subsubsection{High-Capacity Image Steganography}
With the advent and development of deep learning, high-capacity image steganography appears based on the encoding (concealment) and decoding (restoration) networks~\cite{weng2019high}. 
To leverage the high sensitivity to tiny input perturbations of deep neural networks, Zhu~\textit{et al.}~\cite{zhu2018hidden} hid and restored the secret data with the help of cover images. Baluja~\textit{et al.}~\cite{baluja2017hiding} attempted to place a full-size color image (secret) within another image of the same size (cover) with deep neural networks. Following such direction, Duan~\textit{et al.}~\cite{duan2019reversible} proposed a new image steganography scheme based on a U-Net structure. Similarly, Wu~\textit{et al.}~\cite{wu2018stegnet} and Rahim~\textit{et al.}~\cite{rahim2018end} combined recent deep convolutional neural network methods with image-into-image steganography to hide the same size images successfully. 

These steganographic methods focus on the \textit{security} during transmission but run a risk of the leakage of the secret images after restoration. Our proposed MIAIS framework prevents privacy leakage by performing recognition directly on container images without restoring the secret images. 

\subsection{Adversarial Learning}
\label{Related_Work_Adversarial_Learning}


To enhance anti-analysis ability, many steganography methods adopted GAN-based adversarial learning. The pipelines of these GAN-based methods were similar, but the implementations were different~\cite{tang2017automatic, Volkhonskiy2017SGAN, shi2017ssgan}. 
In order to achieve better invisible steganography, Zhang~\textit{et al.}\cite{zhang2019invisible} implemented a steganalyzer based on the divergence on the empirical probability distributions. To optimize the perceptual quality of the images, Zhang~\textit{et al.}~\cite{zhang2019steganogan} proposed a novel technique named SteganoGAN for hiding arbitrary binary data in images. 
Hayes~\textit{et al.}~\cite{hayes2017generating} showed that adversarial training can produce robust steganographic techniques under an unsupervised training scheme. A similar conclusion was demonstrated in Shi~\textit{et al.}'s work~\cite{shi2019synchronized}.

In a word, the aforementioned GAN analogous methods essentially aimed to approximate the distribution of container images with that of cover image domain by a T/F discriminator. But our task aims to perform recognition directly on container images, which means the discriminator is a classification instead of T/F decision. Inspired by the Maximum Classifier Discrepancy (MCD)~\cite{saito2018maximum} in Unsupervised Domain Adaptation (UDA), we design a minimax optimization. Unlike GAN analogous methods, it pays attention to the class distribution between the secret image domain and the container image domain, not just the domain gap between the cover images and container images.

%% file: sections/method.tex
\section{Multitask Identity-Aware Image Steganography}
\begin{table}[tb]
	\centering
	\caption{Notations}
	\label{tab:notation}
	\vspace{-1em}
	\begin{tabu} to 0.49\textwidth {X[c]X[4]}
		\toprule
		$\mathcal{S}$  & a secret image dataset\\
		$\mathcal{C}$  & a set of cover images\\
		$x^s$    & a secret image \\
		$x^c$    & a cover image \\
		$x^h$    & a container image \\
		$h(\cdot)$ & the steg-generator \\
		$f(\cdot)$ & the steg-classifier without minimax optimization\\
		$f_1(\cdot)$, $f_2(\cdot)$ & the two steg-classifiers in the minimax optimization\\
		$r(\cdot)$ & the steg-restorer \\
		$\theta_h$ & the parameters of $h(\cdot)$ \\
		$\theta_f$ & the parameters of $f(\cdot)$ \\
		$\theta_{f_1}$ & the parameters of $f_1(\cdot)$ \\
		$\theta_{f_2}$ & the parameters of $f_2(\cdot)$ \\
		$\theta_r$ & the parameters of $r(\cdot)$ \\
		\midrule
		$\mathcal{L}_{cont}$    & the content loss \\
		$\mathcal{L}_{vis}$    & the visual similarity loss \\
		$\mathcal{L}_{id}$    & the recognition loss \\
		$\mathcal{L}_{ce}$    & the standard cross-entropy loss \\
		$\mathcal{L}_{trip}$    & the standard triplet loss \\
		$\mathcal{L}_{res}$    & the restoration loss \\
		$\mathcal{L}_{dis}$ & the discrepancy loss \\
		\bottomrule
	\end{tabu}
\end{table}

\begin{figure*}[htbp]
	\centering
	\makebox[\textwidth][c]{
		\includegraphics[scale=0.50]{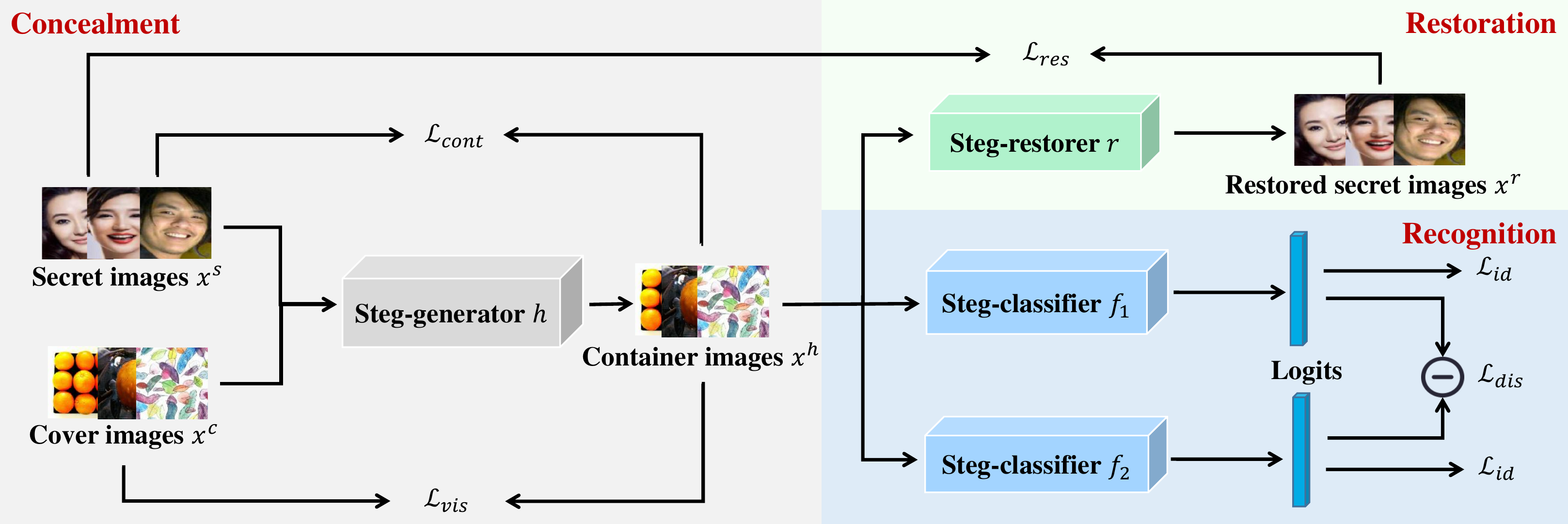}
	}
	\caption{Overview of Multitask Identity-Aware Image Steganography (MIAIS). The framework is composed of a network for hiding secret images called a \textbf{steg-generator}, two classifiers to recognize container images called \textbf{steg-classifiers}, and a network to restore secret images called a \textbf{steg-restorer}.}
	\label{pipeline}
\end{figure*}

Multitask Identity-Aware Image Steganography (MIAIS) consists of a recognition branch and an optional restoration branch, both of which share a concealment part, as shown in \cref{pipeline}. Thus, the receiver can perform direct recognition on the container image with steg-classifier for privacy protection (the single-task scenario), or alternatively, perform the recognition and restoration (the multitask scenario) to make the secret image restoration flexible in some cases.

In this section, we first specify the whole architecture of MIAIS. Second, we illustrate the objective functions and the optimization process. Finally, we design a minimax optimization to deal with the contradictory issues during optimization.
For convenience, \cref{tab:notation} summarizes the notations.


\subsection{Architecture}
\label{IPIS_architecture}

In this part, we walk through the architecture of MIAIS and introduce the data flow.

\subsubsection{The Steg-Generator}
The \textbf{steg-generator} generates a container image to hide a secret image in a cover image. 
Let $\mathcal{S}$ be a secret image dataset and $\mathcal{C}$ be a set of cover images. Given $x^s \in \mathcal{S}$ and $x^c \in \mathcal{C}$, the container image $x^h$ for $x^s$ is 
\begin{equation}
\label{hiding_net}
x^h = h ( {x}^{s}, {x}^c ; \theta_h ) ,
\end{equation}
where $h(\cdot)$ denotes the steg-generator and $\theta_h$ is its parameters.

\subsubsection{The Steg-Classifiers}
To prevent the privacy leakage after restoration, we propose a \textbf{steg-classifier} to perform recognition directly on container images. Let $f(\cdot)$ be a steg-classifier and $\theta_f$ be its parameters. The predicted probability $p$ of the container image $x^h$ is 
\begin{equation}
\label{predict_result}
p  =\sigma ( f(x^h; \theta_f ) ) ,
\end{equation}
where $\sigma (\cdot)$ is the softmax function and $p \in [0,1]^K$, where $K$ is the class amount of secret images. We will see how we use two competing steg-classifiers to improve the recognition performance in \cref{ATSAL_Strategy}.

\subsubsection{The Steg-Restorer}
To compare our method with traditional steganography, we embed a restoration branch in our framework. The \textbf{steg-restorer}, denoted by $r\left(\cdot\right)$, inputs $ x^h $ and outputs a restored secret image $x^r$. That is, 
\begin{equation}
x^r= r ( {x}^{h}; \theta_r ),
\end{equation}
where $\theta_r$ is the parameters of $r(\cdot)$.

\subsection{Loss Functions}
\label{problem-formulation}

In this part, we illustrate the loss functions in MIAIS according to the task scenario: concealment, recognition, and restoration. 

\subsubsection{Concealment}
Firstly, our framework aims to hide a given secret image $x^s$ into a cover image $x^c$, producing a container image $x^h$. To make $ x^h $ look similar to $x^c$, a \emph{visual similarity loss} is used to evaluate the consistency, i.e.,
\begin{equation}
\label{cover_contaner_loss}
\mathcal{L}_{vis}\left(h\right) = 1 - \frac{1}{B}\sum_{i=1}^{B} \text{MS-SSIM}( {x}_{i}^{h} , {x}_{i}^{c} ) ,
\end{equation}
where $x_i^h$ and $x_i^c$ are the $i$-th container and cover image in the batch respectively, and $\text{MS-SSIM}$ is \textbf{M}ulti-\textbf{S}cale \textbf{S}tructural \textbf{Sim}ilarity~\cite{wang2003multiscale}, widely used in image steganography~\cite{zhang2019invisible}. 

During the concealment, our framework requires $x^h$ to preserve the identity information of $x^s$, in order to perform recognition directly on container images. To preserve the identity information, we propose a \textbf{content loss}, inspired by the perceptual loss~\cite{johnson2016perceptual} in the area of style transfer~\cite{huang2017arbitrary}. That is, 
\begin{equation}
\label{content_loss}
\mathcal{L}_{cont}(h) = \frac{1}{B}\sum_{i=1}^{B} || v({x}_{i}^{s}; \theta_v^*) - v({x}_{i}^{h}; \theta_v^*) ||_2 ,
\end{equation}
where $B$ is the batch size, $x_i^s$ and $x_i^h$ are the $i$-th secret and container image in the batch respectively, $v(\cdot)$ is the feature extractor of $f(\cdot)$ pre-trained on the secret image dataset $\mathcal{S}$, and $\theta_v^*$ is its parameters, which are  fixed in our framework. The feature extractor is essentially the partial deep neural network of our steg-classifier except for the last fully connected layer.

\subsubsection{Recognition}

To measure the recognition performance, we denote the recognition loss as $\mathcal{L}_{id}\left(f,h\right)$. For the normal image classification task, we adopt the standard cross-entropy loss function as the recognition loss, That is, $\mathcal{L}_{id} \left(f,h\right) = \mathcal{L}_{ce}\left(f,h\right)$,
\begin{equation}
\label{CE_1}
\mathcal{L}_{ce} \left(f,h\right) 
= - \frac{1}{B\cdot K}\sum_{i=1}^{B} \sum_{k=1}^{K} q_i[k] \log({p}_i [k]),
\end{equation}
where $K$ is the class amount of secret images, $p_i$ is the predicted probability of $x_i^s$ and $q_i \in \{0,1\}^K$ is the one-hot ground truth vector of $x_i^s$. 

As for the face verification task, we add an extra triplet loss term into the recognition loss. That is, $\mathcal{L}_{id} \left(f,h\right) = \mathcal{L}_{ce} \left(f,h\right) + \alpha \mathcal{L}_{trip} \left(f,h\right)$,
\begin{align}
\label{trip_loss}
\begin{split}
\mathcal{L}_{trip} \left(f,h\right) &= \sum_{i=1}^{B} \big[ {|| f({x}_{i}^{h,a}) - f({x}_{i}^{h,p}) ||}_2 \\ 
& \qquad \  - {|| f({x}_{i}^{h,a}) - f({x}_{i}^{h,n}) ||}_2 + \gamma \big]_{+},
\end{split}
\end{align}
where ${x}_{i}^{h,a}$, ${x}_{i}^{h,p}$, ${x}_{i}^{h,n}$ represent anchor, positive and negative examples respectively, and $\gamma$ is a margin that is enforced between positive and negative pairs.
\subsubsection{Restoration}
An optional objective of MIAIS is restoring secret images. Akin to $\mathcal{L}_{vis}$, the \textit{restoration loss} is 
\begin{equation}
\label{reveal_loss}
\mathcal{L}_{res}\left(h,r\right) = 1 - \frac{1}{B}\sum_{i=1}^{B} \text{MS-SSIM}( {x}_{i}^{s} , {x}_{i}^{r} ) .
\end{equation}

\subsection{Minimax Optimization}
\label{ATSAL_Strategy}

\begin{figure*}[htbp]
	\centering
	\makebox[\textwidth][c]{
		\includegraphics[scale=0.36]{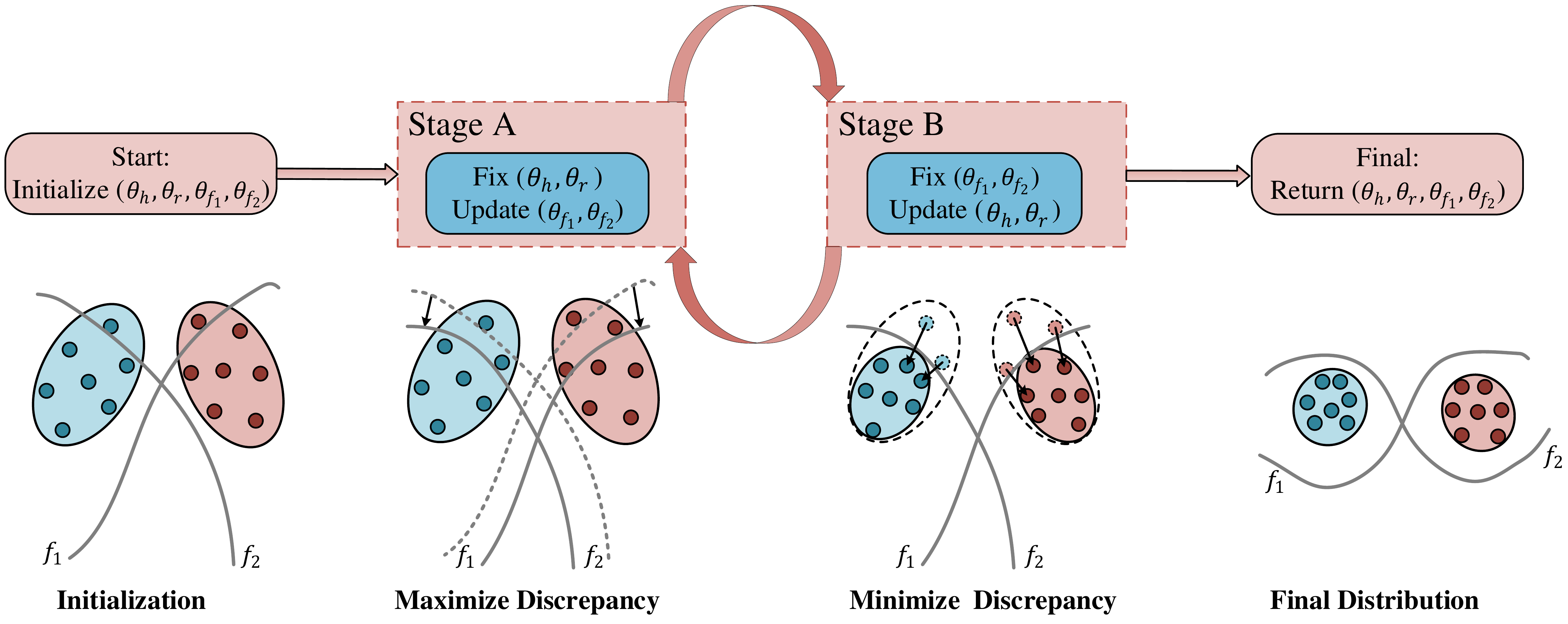}
	}
	\caption{ Illustration of the minimax optimization.
		 The optimization consists of two stages. At \textbf{Stage A}, with the fixed steg-generator (container images), to \textbf{maximize the discrepancy}, the two decision boundaries $f_1(\cdot)$ and $f_2(\cdot)$ are optimized for the farther distances, resulting in narrower intersection region. At \textbf{Stage B}, with the two steg-classifiers fixed, \textbf{minimizing the discrepancy} means to move the features of the container images towards the intersection region of these two steg-classifier decision boundaries $f_1(\cdot)$ and $f_2(\cdot)$, through training the steg-generator. After repeating the above two stages, the intra-class distance is reduced and inter-class distance is increased, as shown in the \textbf{final distribution}. Best viewed in color. Note that for the \textbf{single-task} scenario, only the $\theta_h$ is optimized in the minimax optimization. For the \textbf{multitask} scenario, $\theta_h$ and $\theta_r$ are jointly optimized in the minimax optimization. 
	}
	\label{Fig_ATSAL}
\end{figure*}

\subsubsection{Objective}

In the single-task scenario (recognition), the overall objective is to optimize the steg-generator and the steg-classifier by minimizing the summation of visual similarity loss, the classification loss, and the content loss, i.e.,
\begin{equation}
\label{argmin_option1_WO_recovery}
\theta_h^*, \theta_f^* = \argmin_{\theta_h, \theta_f}  \mathcal{L}_{vis}\left(h\right) + 
\mathcal{L}_{id} \left(f,h\right) + \mathcal{L}_{cont}(h).
\end{equation}

In the multitask scenario (joint recognition and restoration), the overall objective is to optimize the steg-generator, the steg-classifier, and the steg-restorer by minimizing the summation of visual similarity loss, the classification loss, the content loss, and the restoration loss. That is,
\begin{equation}
\begin{aligned}
\label{argmin_option2_W_recovery}
\theta_h^*, \theta_f^*,\theta_r^* = \argmin_{\theta_h, \theta_f,\theta_r} 
& \; \mathcal{L}_{vis}\left(h\right) +
\mathcal{L}_{id} \left(f,h\right)\\
& \; + \mathcal{L}_{cont}(h) + \mathcal{L}_{res}\left(h,r\right) .
\end{aligned}
\end{equation}


During the optimization, the parameters of steg-classifier $\theta_f$ are updated by the gradients of $\mathcal{L}_{id}$. But the input of steg-classifier, the container image ${x}^{h}$, depends on the $\theta_h$, which accounts for the calculation of $\mathcal{L}_{id}$ depends not only on $\theta_f$, but also on $\theta_h$. The performance of concealment depends on $\theta_h$, while that of recognition depends on $\theta_f$. Therefore, it is hard to seek for a trade-off between the concealment and recognition when optimizing the steg-generator and steg-classifier simultaneously. To address this issue, we adopt the minimax optimization to iteratively update each of the two parts with the other one fixed, as shown in \mbox{\cref{Fig_ATSAL}}.

To design a minimax optimization, we introduce two steg-classifiers, denoted by $f_1(\cdot)$ and $f_2(\cdot)$. After initialization, we train our framework in two alternate stages iteratively. At \textbf{Stage A}, $f_1(\cdot)$ and $f_2(\cdot)$ are trained to maximize the discrepancy between them with $h(\cdot)$ fixed. At \textbf{Stage B}, the steg-generator $h(\cdot)$ is trained to minimize the discrepancy between the fixed $f_1(\cdot)$ and $f_2(\cdot)$. The process is shown in Algorithm~\ref{AAPP_algorithm}. The details of the discrepancy loss and the two stages are described as follows.

\begin{algorithm}[!htbp]
	\caption{The minimax optimization}
	\label{AAPP_algorithm}
	\hspace*{0.02in} {\bf Input:} 
	Secret image dataset $\mathcal{S}$; \\ 
	\hspace*{0.45in} Cover image set $\mathcal{C}$;  
	\begin{algorithmic}[1]
		\STATE Initialize (parameters $\theta_h, \theta_r, \theta_{f_1}, \theta_{f_2}, \theta_v^* $);\\ 
		\FOR {epoch $ = 1, \cdots$, \#epochs}
		\FOR {every batch}
		\STATE {Load and normalize $B$ (the batch size) secret images from $\mathcal{S}$;}\\
		\STATE {Randomly sample $B$ cover images from $\mathcal{C}$;}\\
		\STATE {Generate $B$ container images and $B$ restored images;}\\
		\IF {epoch \% 2 == 1}
		
		\STATE{\textbf{Stage A:}}
		\STATE {Update $\theta_{f_1}, \theta_{f_2}$ (\cref{argmin_stage_B});}\\
		\ELSE
		\STATE{\textbf{Stage B:}}
		\STATE {Update $\theta_h$ for the recognition\\
			(and update $\theta_r$ in the multitask scenario)\\ (\cref{stage_A_single_task} or \cref{stage_A_multi_task});}\\
		\ENDIF
		\ENDFOR
		\ENDFOR
	\end{algorithmic}
	\hspace*{0.02in} {\bf Output:} 
	$\theta_h, \theta_r, \theta_{f_1}, \theta_{f_2}$
	
\end{algorithm} 

\begin{figure*}
	\centering
	\makebox[\textwidth][c]{\includegraphics[scale=0.56]{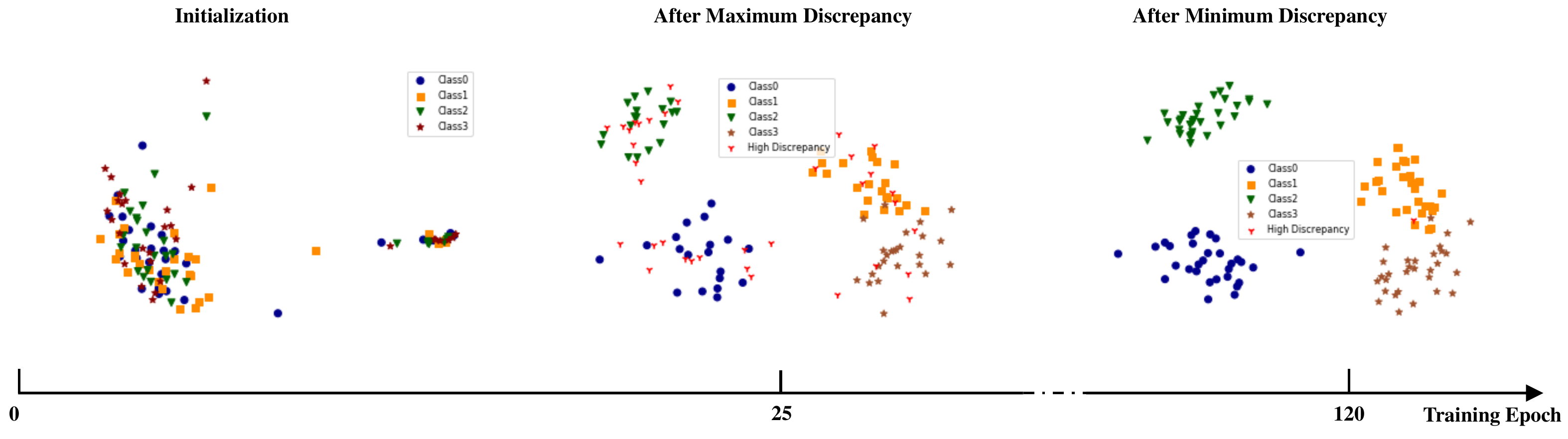}}
	\caption{T-SNE visualization of the feature distribution of four classes at different stages on the AFD dataset. Each class has 30 samples. We extract the initial container features before training, and then extract the container features after maximum discrepancy and minimum discrepancy in the 25th epoch and the 120th epoch, respectively. To clearly show the samples with large discrepancies, we highlight the samples with a discrepancy more than 3e-4 by special red marker (tri-down). } 
	\label{minimax_tsne}
\end{figure*}

\subsubsection{Discrepancy Loss} Given an arbitrary container image $x^h$, $f_{1}(x^h; \theta_{f_1})$ and $f_{2}(x^h; \theta_{f_2})$ denote the logits output of two steg-classifiers respectively. Similar to many minimax optimization method~\cite{saito2017maximum,lee2019drop}, to measure the differences between the prediction results of the two steg-classifiers, we define the discrepancy loss based on absolute values of the difference between the two logits:
\begin{equation}
\label{dis_loss}
\mathcal{L}_{dis}(f_1, f_2) = \frac{1}{B\cdot K}\sum_{i=1}^{B} \sum_{k=1}^{K} | f_{1}(x_i^h; \theta_{f_1})[k] - f_{2}(x_i^h; \theta_{f_2})[k] | .
\end{equation}

\subsubsection{Training}
We train our framework with $\mathcal{L}_{dis}$ in a two-stage manner.

\textbf{Stage A} \textit{(Maximize Discrepancy)}: At this stage, we aim to maximize the discrepancy between the two steg-classifiers. With the steg-generator $h(\cdot)$ fixed, we train the two steg-classifiers $f_1(\cdot)$ and $f_2(\cdot)$ to minimize the container classification loss $\mathcal{L}_{id}$ and maximize the discrepancy loss $\mathcal{L}_{dis}$, i.e.,
\begin{equation}
\label{argmin_stage_B}
\theta_{f_1}^*, \theta_{f_2}^* = \argmin_{\theta_{f_1}, \theta_{f_2}} \mathcal{L}_{id}(f_1, h) + \mathcal{L}_{id}(f_2, h) - \mathcal{L}_{dis}(f_1, f_2) .
\end{equation}

Maximizing the two steg-classifiers discrepancy leads to two far-distant class decision boundaries derived from the two steg-classifiers. Features within the intersection region of these two boundaries are effective for classification because these features are discriminative to any classifier whose decision boundary is between the two far-distant decision boundaries. Then at the next \textbf{Stage B}, with the two steg-classifiers fixed, the minimization of the discrepancy causes the features of the container images to move towards the intersection region of these two steg-classifier decision boundaries by training the steg-generator, as shown in the bottom of \cref{Fig_ATSAL}. 

\textbf{Stage B} \textit{(Minimize Discrepancy)}: At this stage, we aim to minimize the discrepancy between the two steg-classifiers. We train the steg-generator $h(\cdot)$ to minimize the discrepancy loss $\mathcal{L}_{dis}$ of two fixed steg-classifiers $f_1(\cdot)$ and $f_2(\cdot)$, visual similarity loss $\mathcal{L}_{vis}$, and content loss $\mathcal{L}_{cont}$ jointly. 
Then, the optimization objective in \cref{argmin_option1_WO_recovery} for the single-task scenario is rewritten as follows:
\begin{equation}
\label{stage_A_single_task}
\begin{aligned}
\theta_h^* = \argmin_{\theta_h}  
& \; \mathcal{L}_{vis}\left(h\right) + \mathcal{L}_{id} \left(f_1,h\right) + \mathcal{L}_{id} \left(f_2,h\right) \\
& \; + \mathcal{L}_{cont}(h) + \mathcal{L}_{dis}(f_1, f_2).
\end{aligned}
\end{equation}
Note that since the two steg-classifiers are fixed, $\theta_{f_1}$ and $\theta_{f_2}$ do not need to be optimized. In the single-task scenario, only the $\theta_h$ are optimized in minimize discrepancy stage.


As for the multitask scenario, we just need to add the restoration loss $\mathcal{L}_{res}$ to \cref{stage_A_single_task} and replace the optimization objective of \textbf{Stage B} as follows: 
\begin{equation}
\label{stage_A_multi_task}
\begin{aligned}
\theta_h^*, \theta_r^* =
& \;  \argmin_{\theta_h, \theta_r} \mathcal{L}_{vis}(h) + \mathcal{L}_{id} \left(f_1,h\right) + \mathcal{L}_{id} \left(f_2,h\right) \\
& \; + \mathcal{L}_{cont}(h) + \mathcal{L}_{dis}(f_1, f_2) + \mathcal{L}_{res}(h,r).
\end{aligned}
\end{equation}

In the multitask scenario, $\theta_h$ and $\theta_r$ are jointly optimized in minimize discrepancy stage.

We further visualize the intermediate snapshots of container image features during the minimax process using T-SNE. As shown in \mbox{\cref{minimax_tsne}}, after maximum discrepancy, the samples with large discrepancy (red markers) increase, which means the discrepancy of the two steg-classifiers enlarges. After the minimum discrepancy in \mbox{\cref{minimax_tsne}}, the red markers decrease, which indicates the classification difference of the samples is significantly reduced. After this minimax discrepancy process, the intra-class distance is reduced and the inter-class distance is increased, resulting in the compact and discriminative features of container images for recognition. The coefficients to balance the weight among the loss terms in \cref{stage_A_single_task,stage_A_multi_task,argmin_stage_B} will be clarified \textit{in Section I of the supplemental material} for reproduction.


%% file: sections/experiment.tex
\section{Experiments}

In this section, we first describe the experimental setup. Then, we perform extensive experiments on the direct recognition task. Finally, we evaluate the multitask scenario (recognition and restoration) to compare our method with traditional image steganography methods.

\subsection{Experimental Setup}
\label{Experimental_setup}

\subsubsection{Datasets}
We conduct the experiments in two recognition tasks: \emph{face recognition} and \emph{image classification}. 
For \emph{face recognition}, we conduct the experiments on the \textbf{AFD} (Asian Face Dataset)~\cite{xiong2018asian} dataset.  We use 310,969 images of 1,662 identities for training, and 35,780 images of 356 identities for validation. 
For \emph{image classification}, we use the \textbf{Tiny-Imagenet} dataset, consisting of 200 classes, each of which has 500 training, 50 validation, and 50 test images. 

For the cover image selection in steganography, we use 100 images from the style transfer and satellite datasets, which have a rich texture for concealing the secret image information. Note that all the images are unrelated to the training datasets, such as \textbf{AFD} and \textbf{Tiny-ImageNet}. To show the cover$/$carrier robustness of our proposed methods, we use \textbf{LFW}, \textbf{Pascal VOC 2012}, and \textbf{ImageNet} as the cover image set respectively in the single-task scenario. \textbf{Pascal VOC 2012} is a dataset designed for object detection and semantic segmentation and contains 33260 images.  \textbf{LFW} (Labeled Faces in the Wild) dataset contains 13233 images of 5749 identities. \textbf{ImageNet} is a large visual database in object recognition.

\subsubsection{Network Architecture} Similar to many image steganography methods\mbox{\cite{weng2019high,duan2019reversible}}, the \textbf{steg-generator} is U-net\mbox{\cite{ronneberger2015u_net}} structure. Both \textbf{steg-classifiers} are the Inception network structure\mbox{\cite{szegedy2016rethinking}}. We design a simple \textbf{steg-restorer} inspired by the similar architecture\mbox{\cite{weng2019high}}. The channels of input and output are shown in \mbox{\cref{reconstruction}}. Each layer of the \textbf{steg-restorer} has a $3\times 3$ convolution one-padding (a stride of 1). There is a batch normalization (BN) layer and a rectified linear unit (ReLU) after each convolution layer except for the last one. The results of the last convolution layer will be activated by a Sigmoid function, then output as the restored secret image. We draw the network architecture details for reproduction. \textit{Please see Section IX of the supplemental material for more details.}

\input{tables/reconstruction_arch}

\subsubsection{Evaluation Metrics}
From the perspective of privacy protection, we measure the image perturbations between the cover and container image using Mean Square Error (\textbf{MSE}), Multi-Scale Structural Similarity (\textbf{MS-SSIM})~\cite{wang2004image,ma2016group}, and Peak Signal to Noise Ratio (\textbf{PSNR}). The Detection Rate (\textbf{DR}) is the result of the StegExpose\mbox{\cite{boehm2014stegexpose}} which is a popular statistical steganalysis toolkit. Our main focus, the recognition accuracy (\textbf{Acc}) on container images denotes the classification accuracy in the image classification task or identification accuracy in the face recognition task. To evaluate the MIAIS framework embedded with the restoration module quantitatively, we use the \textbf{MSE} and \textbf{MS-SSIM} to measure the distortion between secret and restored secret images. 

\subsubsection{Implementation Details}
All experiments are performed on a workstation with two Titan X GPU cards. The input images are rescaled to $256 \times 256$. In all of our experiments, we use an SGD optimizer to update the parameters of our networks. The batch size is set to 16 for face recognition, 32 for image classification in the single-task scenario, and 4 for face restoration in the multitask scenario. The learning rate is initialized by 0.001 in the single-task scenario and 0.0001 in the multitask scenario, and reduced by a factor of 0.2 when the overall loss stagnates. In order to evaluate the effectiveness of identity preserving at a certain level of visual information hiding, we set a hiding threshold to control the visual quality of steganography. We dynamically adjust the weights of losses in \cref{stage_A_single_task,stage_A_multi_task} through the magnitude relation between $\mathcal{L}_{vis}$ and the hiding threshold. In order to determine the hiding threshold, we train our baseline for large iterations and observe that visual loss (MS-SSIM) converges to about 0.02 on the test dataset (\textit{Please see Section II of the supplemental material for more details}). Therefore, we set the hiding threshold to 0.02 in our experiments. As described in the \mbox{\cref{problem-formulation}}, we adopt the triplet loss\mbox{\cite{schroff2015facenet}} for face recognition to obtain the better performance ($\alpha = 2$, $\gamma = 1.2$).

\subsection{Experiments in Direct Recognition}
\label{Experiments_Recognition}


\input{tables/ablation}

\begin{figure*}
	\centering
	\includegraphics[width=.9\textwidth]{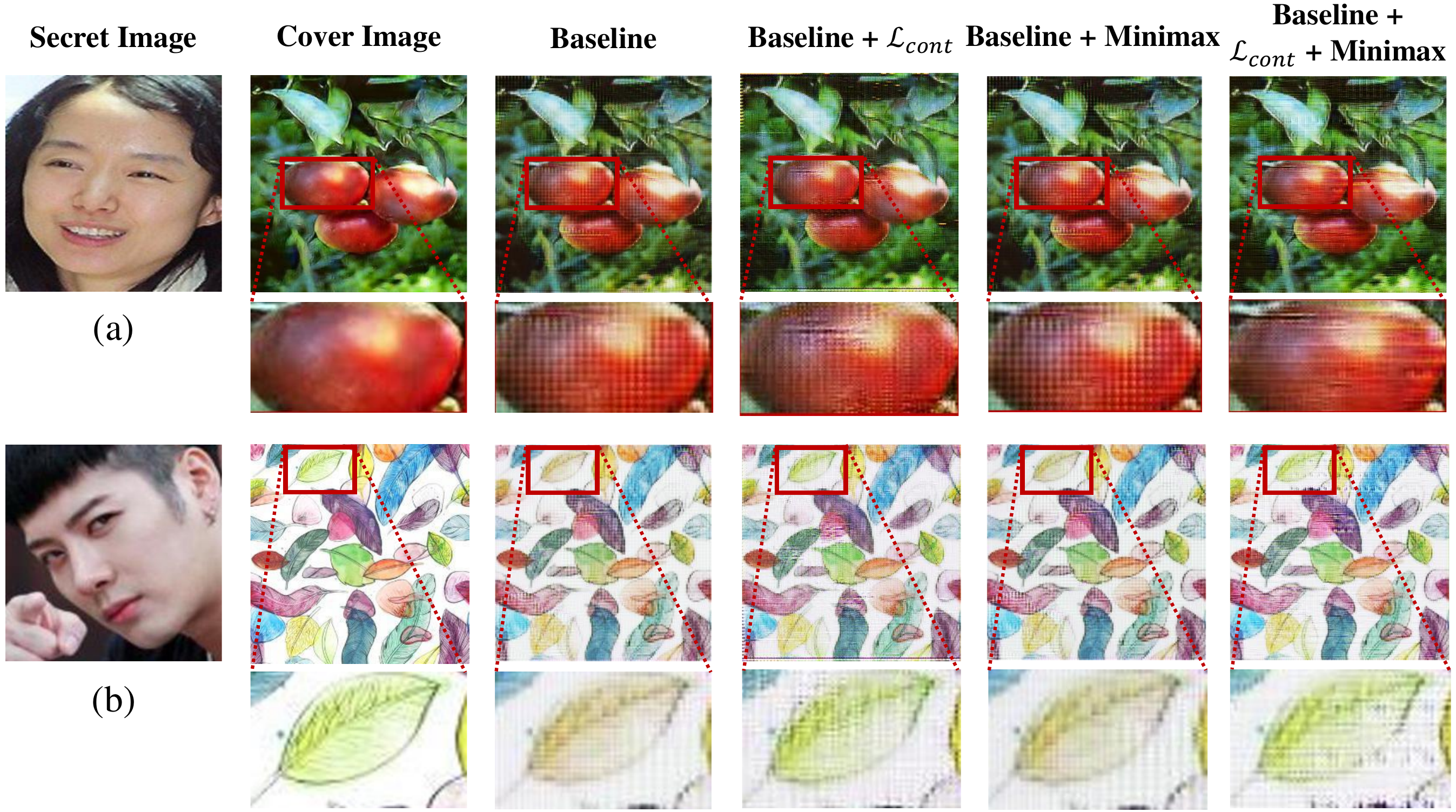}
	\caption{Visual quality of ablative models on the AFD dataset. 
	The first two columns show secret images and cover images. The rest columns show the container images produced by ablative models. The ablative method \textbf{Baseline + $\mathcal{L}_{cont}$} is more colorful and vivid compared with the method \textbf{Baseline}. The methods without content loss introduce more mosaics and lattices. However, the content loss would bring about some artifacts, which affects the visual quality. Best viewed in color.
	}
	\label{ablation_sectionB}
\end{figure*}

\subsubsection{Ablation Study}
To perform a series of ablation experiments, we choose the images in AFD or Tiny-ImageNet as the secret images. We calculate the MSE, PSNR, and MS-SSIM between the cover and the container images for quantitative evaluation. In order to measure the secrecy of steganography, we show the detection rate (DR) of StegExpose in the container image domain. The comparison methods of ablation experiments are shown as follows:

\begin{itemize}
	\setlength{\itemsep}{0pt}
	\setlength{\parsep}{0pt}
	\setlength{\parskip}{0pt}
	
	\item \textbf{Baseline.} We train one steg-generator paired with one steg-classifier with the summation of $\mathcal{L}_{vis}$ and $\mathcal{L}_{ce}$ as our baseline.
	\item \textbf{Baseline + $\mathcal{L}_{cont}$.} We add the $\mathcal{L}_{cont}$ to baseline and train the steg-generator together with the steg-classifier jointly (\cref{argmin_option1_WO_recovery}).
	\item \textbf{Baseline + Minimax.} We train one steg-generator paired with two steg-classifiers via the minimax optimization.
	\item \textbf{Baseline + $\mathcal{L}_{cont}$ + Minimax.} We train one steg-generator paired with two steg-classifiers with $\mathcal{L}_{cont}$ via the minimax optimization.
\end{itemize}

\emph{Quantitative Evaluation.}
As shown in Table~\ref{Table_Ablation_AFD_TI}, in both the face recognition and image classification tasks, the methods with $\mathcal{L}_{cont}$ are better than the methods without $\mathcal{L}_{cont}$ in Acc, which indicates that the similarity constraint between the secret images and container images preserve the identity (class) information. 
Most importantly, the method using the minimax optimization performs better than the baseline (joint training of single steg-classifier) over 2\%, which shows that our minimax optimization obtains more discriminative features of the container images. As defined in \cref{cover_contaner_loss}, $\mathcal{L}_{vis}$ is related to MS-SSIM. Therefore, in \cref{Table_Ablation_AFD_TI}, the MS-SSIM is similar through the hiding threshold control. Within a similar visual loss range, the detection rates of all methods are not far from random guessing. Under the similar visual quality of hiding measured by MS-SSIM between the container image and the cover image, the methods with the minimax optimization are better than the methods without it. From \mbox{\cref{Table_Ablation_AFD_TI}}, the \textbf{Baseline} method has higher PSNR and MS-SSIM than the \textbf{Baseline + $\mathcal{L}_{cont}$} method, which means that the method without the content loss can preserve more appearance information and has the better visual quality. In fact, the content loss acts on the constraint of the global discriminative identity features, which is indeed detrimental to the visual quality of images.

\emph{Visual Quality.}
Through the control of the hiding threshold for visual loss, all the generated images have an imperceptible perturbation compared with the original cover image. But it is interesting that, although every method has an almost similar MS-SSIM, their generated images have different styles. 
As shown in \mbox{\cref{ablation_sectionB}}, the ablative method \textbf{Baseline + $\mathcal{L}_{cont}$} is more colorful and vivid compared with the method \textbf{Baseline}. The methods without content loss introduce more mosaics and lattices. However, the content loss would bring about some artifacts, which affects the visual quality. For example, as highlighted in the red box of the first example (a), we can see the methods without content loss has unrealistic grids compared with the methods with content loss. In the second example (b), the color of green leaves highlighted in the red boxes is preserved well by the ablative models with content loss. \textit{Please see Section III of the supplemental material for the visual quality and color histograms of more examples.}

\input{tables/robust_cover}

\subsubsection{Robustness for the Cover Sets} \label{robustness}
In practical applications, the cover sets are various due to different communication scenarios. Therefore, it is necessary to evaluate the robustness of the results across different datasets. We choose three different task datasets (LFW, Pascal VOC 2012, and ImageNet) as the cover set separately and use the vanilla transfer learning pipeline --- fine-tuning to show the carrier robustness of our MIAIS framework. The model is pre-trained with 100 cover images from the style transfer and satellite datasets. From \mbox{\cref{Table_cover_robustness}}, the recognition accuracy fluctuates during 0.05\%~0.28\% on AFD dataset, and 0.04\%~0.30\% on Tiny-ImageNet dataset. The MSE, MS-SSIM, and PSNR fluctuate within 0.06\%, 0.16\%, 0.60 separately on average, which demonstrates that the results can be transferred across datasets.

The scope of our work is to achieve a trade-off between direct recognition measured by the accuracy and visual concealment measured by PSNR. In the experiments, recognition accuracy was our first priority, which possibly sacrificed the PSNR performance to a certain extent. Actually, the visual quality is controlled by the hiding threshold (default setting 0.02). To raise the PSNR result, we adjust the hiding threshold to 0.001 and report our results with high PSNR. \textit{Please see Section V of the supplemental material for more details.}


\subsubsection{Comparison with Others}
To evaluate the effectiveness, we compare the proposed MIAIS framework with some other visual information hiding methods on AFD datasets. Take the face recognition result on \emph{secret image} as upper bound, the competing methods include Combined Cat Map (CCM)~\cite{wang2015novel}, Learnable Encryption (LE)~\cite{tanaka2018learnable}, PBIE~\cite{sirichotedumrong2019pixel}, Encryption-Then-Compression (EtC)~\cite{kawamura2020privacy}, Extended Learnable Encryption (ELE)~\cite{madono2020block}, EM-HOG~\cite{kitayama2019hog}, Defeating Image Obfuscation (DIO)~\cite{mcpherson2016defeating}, Privacy Preserving Face Recognition (PPFR)~\cite{ergun2014privacy} and a naive Block-Wise pixel Shuffle (BWS). BWS only naively shuffles the $4 \times 4$ pixel blocks without consideration for privacy protection. In Table~\ref{table_comparison_with_others}, we show the MSE, MS-SSIM, PSNR between the secret images and the container images, and accuracy (Acc) of the same steg-classifier model learned on their hiding images. As shown in Table~\ref{table_comparison_with_others}, our proposed MIAIS framework gives the best face recognition performance, which drops only 0.28\% from the upper bound. \textit{Please see Section IV of the supplemental material for the visual quality of these methods.}

\input{tables/comp_classification_other}

\begin{figure}
	\centering
	\includegraphics[width=.36\textheight]{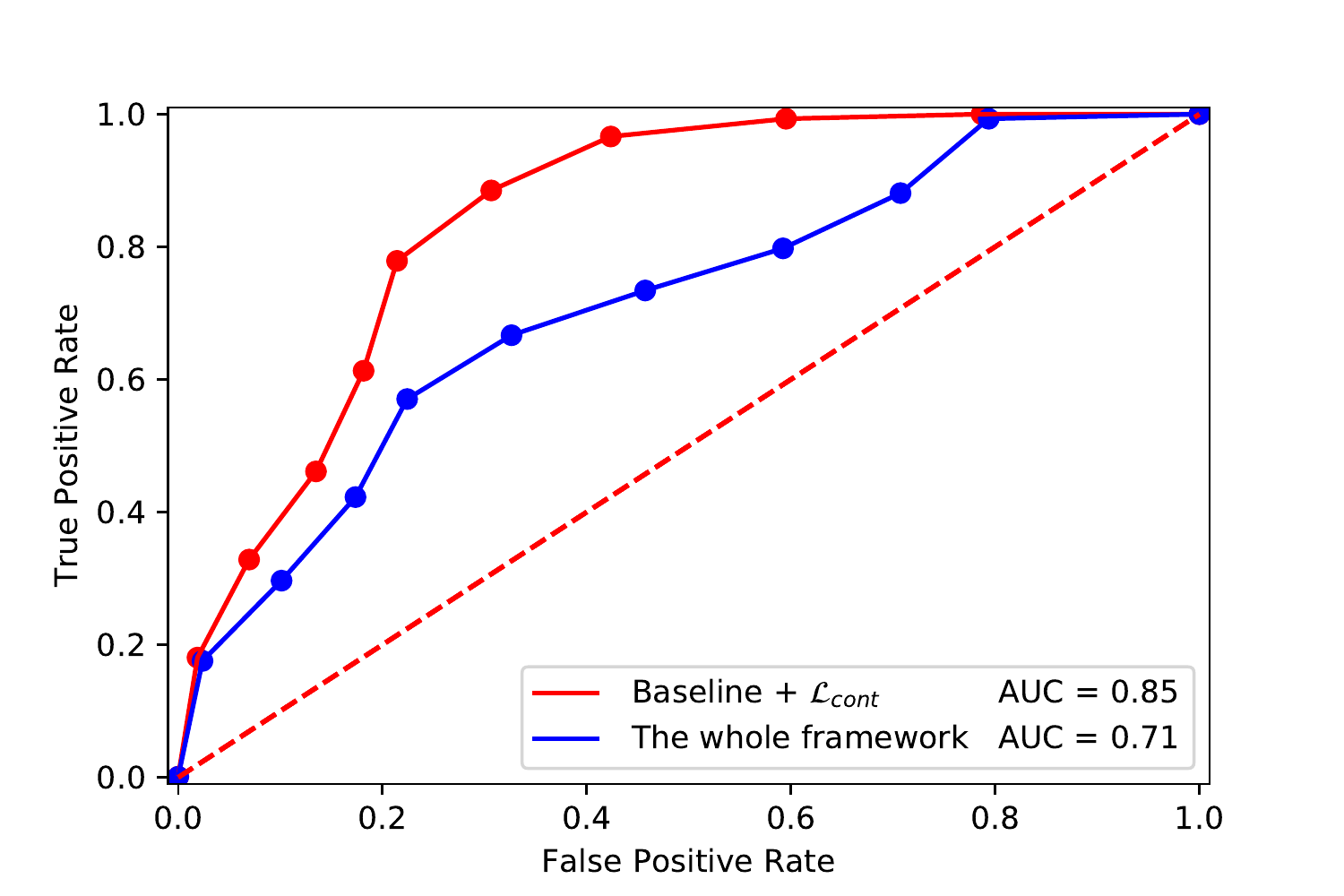}
	\caption{The receiver operating characteristic (ROC) curve produced by the StegExpose for a set of 1000 cover-container pairs.}
	\label{stegexpose}
\end{figure}

\subsubsection{Security Analysis} 
The security of steganography is crucial for privacy protection, which determines the practicability of steganography. We use the popular statistical steganalysis toolkit called StegExpose\mbox{\cite{boehm2014stegexpose}} to analyze our MIAIS framework and the ablation method (\textbf{baseline + $\mathcal{L}_{cont}$}). StegExpose combines several popular steganalysis techniques, such as Sample Pair Analysis~\cite{dumitrescu2002detection}, LSB Steganography Analysis~\cite{fridrich2001reliable,dumitrescu2002steganalysis}, Visual \& Statistical Attacks~\cite{westfeld1999attacks}. The source code of StegExpose is available on Github\footnote{https://github.com/b3dk7/StegExpose.}. Digging into the code, the StegExpose will return a length of the hidden message through the function ``Fuse.se(stegExposeInput)''. The length divided by the total length as a modification score which was ranged from 0 to 1. We follow the experiment setting of HIiPS\mbox{\cite{baluja2017hiding}} to calculate the receiver operating characteristic (ROC) curve.

In the experiment, we select 1000 images randomly from the AFD dataset as the secret image and produce the corresponding 1000 container images through each method. As shown in \cref{stegexpose}, we plot the ROC curve of steganalysis and calculate the area under the curve (AUC). From \cref{stegexpose}, compared with the \textbf{Baseline + $\mathcal{L}_{cont}$}, our MIAIS framework shows the greater ability to fooling the steganalyser (StegExpose), which indicates that the container image can't be discriminated from the cover image by steganalysers. 

\subsection{Experiments in a Multitask Scenario} 
\label{Experiments_MTL}

\input{tables/comp_steg_other}

In this subsection, we conduct our MIAIS framework embedded with the restoration module and other competitive high-capacity image steganography methods under the multitask scenario to jointly restore the secret images and recognize the container images. Similar to experiments on recognition in Section~\ref{Experiments_Recognition}, we calculate MSE, PSNR, and MS-SSIM for cover-container pairs and secret-restored pairs. We calculate DR and accuracy (Acc) for container images. Different from the experiments in recognition, we restore the secret image from the container image and preserve the identity information into container images at the same time. We first introduce the modified steganography methods, then compare these methods to our method from the quantitative and qualitative evaluation. Finally, we perform the security analysis with 3 learning-based steganalysers and the visualization about the spatial and channel modification in an image.

\begin{figure*}
	\centering
	\makebox[\columnwidth][c]{
		\includegraphics[width=.65\textheight]{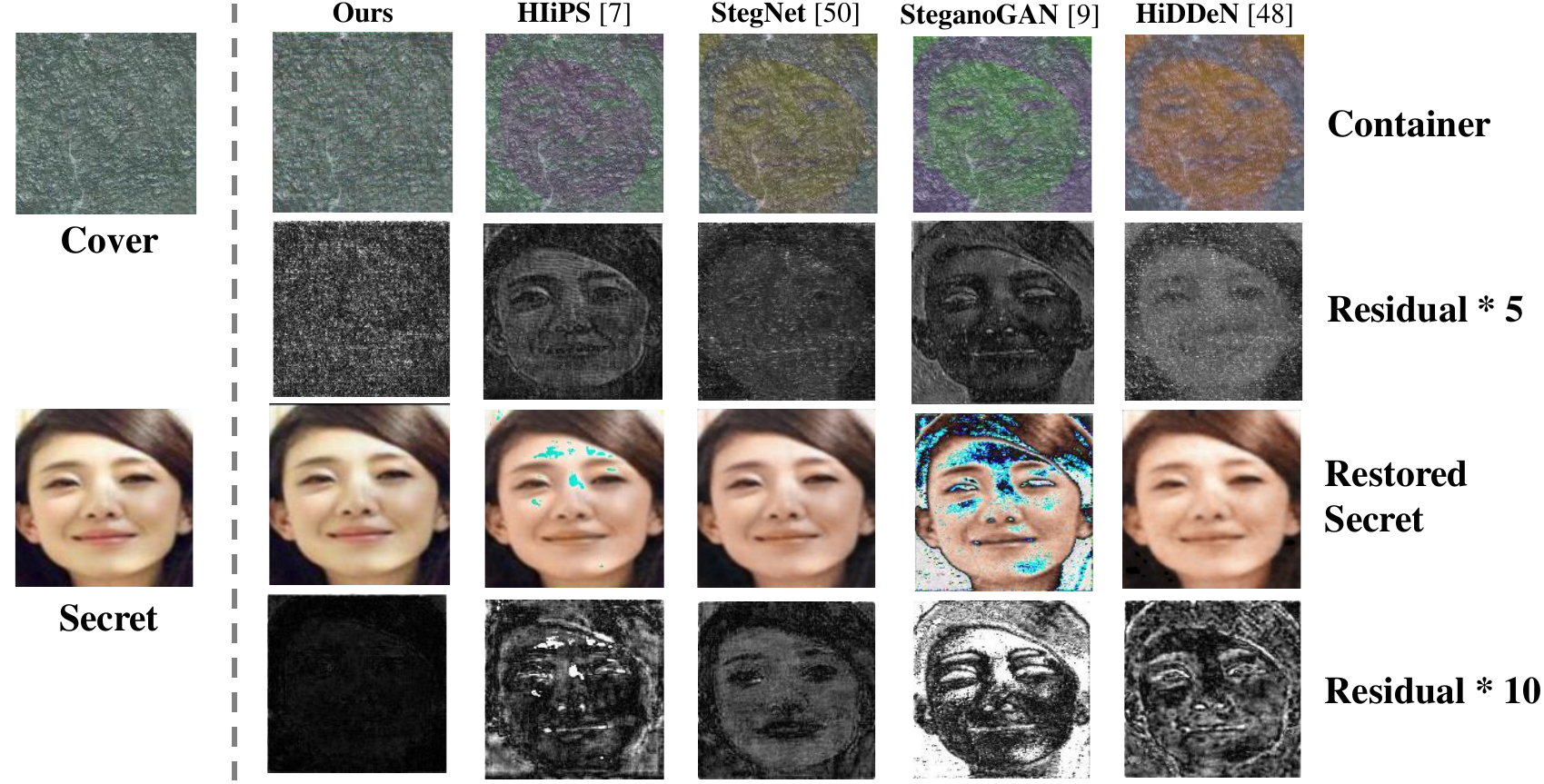}
	}
	\caption{Visual quality of different image steganography methods on the AFD dataset. 
	To show the difference between the cover images and the container images, we enlarge the residual gray images by a factor of 5 or 10. The container images and the restored images produced by our MIAIS look most similar to the cover images and the secret images visually in terms of tones, colors, and sharpness, which can also be seen in the magnified residuals.}
	\label{exp_multi_task_visual}
\end{figure*}

\subsubsection{Competing Methods} We choose for comparison state-of-the-art image steganography methods including HIiPS~\cite{baluja2017hiding}, SteganoGAN~\cite{zhang2019steganogan}, HiDDeN~\cite{zhu2018hidden} and StegNet~\cite{wu2018stegnet}. As HiDDeN~\cite{zhu2018hidden} is not specially designed for high-capacity image steganography, we re-implement its input and output layer to ensure the consistency of experiment settings. The competing methods didn't perform direct recognition on container images. For a fair comparison, we re-implement the aforementioned image steganography methods under the multitask scenario. Specifically, we add the cross-entropy and triplet loss to the other methods and adopt joint training. The modified methods are marked with an asterisk (*), as shown in Table~\ref{Table_Comparison_Steganography}.

\subsubsection{Quantitative Evaluation} As for the performance of concealment/restoration, we focus on the metrics on the cover-container/secret-restored pairs. From Table~\ref{Table_Comparison_Steganography}, our proposed MIAIS framework performs better than the other image steganography methods. Particularly, our framework outperforms the other image steganography methods about 7\% MS-SSIM on average in concealment. As for the performance of recognition, our framework achieves the best results on the Acc, which shows that our method could preserve the identity information more effectively and obtain the discriminative container image for recognition. 
To explore the real-world biometric recognition scenario, we reduce the number of training images by sampling 10 faces per identity. We evaluate the performance of our method and report the results \textit{in the Section VI of the supplemental material}.

As discussed in Section \ref{robustness}, the scope of our work is to achieve a trade-off between direct recognition measured by the accuracy and visual concealment measured by PSNR. Similarly, we report our experimental results with high PSNR under the multi-task scenario. \textit{Please see Section V of the supplemental material for more details.}

\subsubsection{Visual Quality} 
\cref{exp_multi_task_visual} shows the visual quality of different image steganography methods and their corresponding enlarged residual images. The restored secret images produced by our MIAIS are the most similar to the original secret image, while those of the other methods have discordant color areas and blur areas (face). As shown by the residuals of even rows, our MIAIS hides the profile and details of the face very well. These experiments show that our MIAIS framework performs well in the multitask scenario whether from the quantitative or qualitative evaluation. \textit{Please see Section VIII of the supplemental material for more examples.}

\subsubsection{Deep Steganalysis}
Except for the statistical steganalysis --- StegExpose, we further examine our method with learning-based steganalysers\mbox{\cite{ye2017deep,yedroudj2018yedroudj}}. In the experiment, we consider two different cases of whether the steganography model can be available. As shown in \mbox{\cref{deep_steganalysis}}, in the first case, the learning-based steganalysers achieve 100\% detection rate ability over all the steganography methods, as other related work has shown\mbox{\cite{baluja2019hiding,zhang2020udh}}. As for the second case, our method achieves a comparable detection rate compared with the most secure method.

\input{tables/deep_steganalysis}

Similar to other steganography methods, our primary objective is the secret communication without revealing the transported information to a third-party. In our scenario, the transported information is the content/identity of the secret image. Even if a third-party could detect the container image, they cannot recover the content/identity of the secret image. In fact, in real-world applications, the steganography model of our method is confidential and not available, thus ensuring security in terms of resisting detection by the learning-based steganalysis.

\subsubsection{Modification in an Image}
Our framework directly generates the container image on the whole by the deep learning-based steg-generator instead of modifying the least significant bit (LSB). To further show the modification of our steganography network, we do statistics on the distribution of perturbation for each bit and the distribution of the modification values of pixels respectively. \textit{Please see Section VII of the supplemental material for more examples.}

%% file: tables/reconstruction_arch.tex
\begin{table}[tbp]
	\caption{The channels of the \textbf{Steg-restorer}.}
	\label{reconstruction}
	\makebox[\columnwidth][c]{
	\begin{tabular}{ccccccc}
		\toprule
		\textbf{Index} & 1 & 2 & 3 & 4 & 5 & 6\\
		\midrule
		\textbf{Input} & 3 & 64 & 128 & 256 & 128 & 64\\
		\midrule
		\textbf{Output}& 64 & 128 & 256 & 128 & 64 & 3\\
		\bottomrule
	\end{tabular}}
\end{table}

%% file: tables/ablation.tex
\begin{table*}[htbp]
	\caption{Ablation Study of Our Framework. 
	$\uparrow$ means the higher the better; $\downarrow$ means the lower the better.
	}
		\label{Table_Ablation_AFD_TI} 
			\centering
		\begin{tabular}{ccccccccccccc}
			\toprule
			 \multicolumn{3}{c}{\textbf{Methods}} & \multicolumn{5}{c}{\textbf{AFD}} & \multicolumn{5}{c}{\textbf{Tiny-ImageNet}} \\
			\cmidrule(lr){1-3}\cmidrule(lr){4-8}\cmidrule(lr){9-13}
			 \textbf{Baseline} & $\mathcal{L}_{cont}$ & \textbf{Minimax}  
			&  MSE$\downarrow$ & PSNR$\uparrow$ & MS-SSIM$\uparrow$ & DR$\downarrow$ & Acc$\uparrow$ 
			& MSE$\downarrow$ & PSNR$\uparrow$ & MS-SSIM$\uparrow$ & DR$\downarrow$ & Acc$\uparrow$\\
			\toprule
			$\checkmark$ & & & 0.0047  & 23.32  & 0.9808 & 0.521 & 0.8517 
			& 0.0032 & 24.94 & 0.9797 & 0.486 & 0.7335\\ 
			$\checkmark$ & $\checkmark$ &   & 0.0070  & 21.52  & 0.9805  & 0.538  & 0.8648  & 0.0034  & 24.68  &  0.9820  & 0.491 & 0.7353\\ 
			$\checkmark$ & & $\checkmark$ &  0.0058  & 22.37  & 0.9798 & 0.531  & 0.8732 & 0.0032 & 24.98 & 0.9818 & 0.487 & 0.7512  \\ 
			$\checkmark$ & $\checkmark$ & $\checkmark$  &  0.0052  & 22.85  & 0.9799  & 0.586  & 0.8752  & 0.0038 & 24.22 & 0.9819 & 0.497 & 0.7535 \\
			\bottomrule
		\end{tabular}
\end{table*}

%% file: tables/robust_cover.tex
\begin{table*}
	\caption{Robustness on Different Cover Sets.
	$\uparrow$ means the higher the better; $\downarrow$ means the lower the better.
	}
	\centering
		\label{Table_cover_robustness} 
			\begin{tabular}{lcccccccccc}
				\toprule
				\multirow{2}*{\textbf{Cover Set}} & \multicolumn{5}{c}{\textbf{AFD}} & \multicolumn{5}{c}{\textbf{Tiny-ImageNet}} \\
				\cmidrule(lr){2-6}\cmidrule(lr){7-11}
				&  MSE$\downarrow$ & PSNR$\uparrow$ & MS-SSIM$\uparrow$ & DR$\downarrow$ & Acc$\uparrow$ 
				& MSE$\downarrow$ & PSNR$\uparrow$ & MS-SSIM$\uparrow$ & DR$\downarrow$ & Acc$\uparrow$\\
				\toprule
				LFW & 0.0049  & 23.11  & 0.9801 & 0.557 & 0.8730  & 0.0037 & 24.32 & 0.9812 & 0.505 & 0.7545 \\ 
				PASCAL-VOC12  & 0.0057 & 22.45 & 0.9806 & 0.622 & 0.8724  & 0.0042 & 23.77 & 0.9800 & 0.501 & 0.7519\\ 
				ImageNet & 0.0052 & 22.83 & 0.9813 & 0.588 & 0.8729 & 0.0041 & 23.91 & 0.9806 & 0.502 & 0.7515 \\
				\bottomrule
			\end{tabular}
\end{table*}

%% file: tables/comp_classification_other.tex
\begin{table}[htbp]
	\small
	\caption{Comparison with Other Visual Information Hiding Methods on the AFD Dataset.
	$\uparrow$ means the higher the better; $\downarrow$ means the lower the better.
	}
	\label{table_comparison_with_others}
	\makebox[\columnwidth][c]{
		\begin{tabular}{lccccc}
			\toprule
			Method & MSE$\uparrow$ & PSNR$\downarrow$ & MS-SSIM$\downarrow$& Acc $\uparrow$ \\
			\toprule
			Secret Image& 0 & inf & 1 & 0.8780 \\ 
			\toprule
			CCM~\cite{wang2015novel} & 0.1337 & 8.9840 & 0.4179 & 0.5575 \\
			LE~\cite{tanaka2018learnable}  & 0.1437 & 8.5156 & 0.4998 & 0.6154 \\
			PBIE~\cite{sirichotedumrong2019pixel}& 0.1628 & 8.0266 & 0.4106 & 0.5737 \\
			EtC~\cite{kawamura2020privacy}  & 0.1597 & 8.1376 & 0.3811 & 0.5575 \\
			ELE~\cite{madono2020block}  & 0.2951 & 5.5851 & 0.4132 & 0.5830 \\
			EM-HOG~\cite{kitayama2019hog} &  0.1735 & 8.9596 & 0.2279 & 0.5728 \\
			DIO~\cite{mcpherson2016defeating} &  0.0965 & 10.2998 & 0.4646 & 0.8120 \\
			PPFR~\cite{ergun2014privacy}&  0.1260 & 8.9964 & 0.4718 & 0.8540 \\
			BWS & 0.0223 & 17.2696 & 0.9120 & 0.8606 \\
			\midrule
			Ours & 0.1521 & 8.1795 & 0.4191 & 0.8752 \\
			\bottomrule
		\end{tabular}
	}
\end{table}

%% file: tables/comp_steg_other.tex
\begin{table*}
	\centering
	\caption[TABLE]{Comparison with State-of-the-Art High-Capacity Image Steganography Methods on the AFD Dataset.
	$\uparrow$ means the higher the better; $\downarrow$ means the lower the better.
	}
		\label{Table_Comparison_Steganography} 
		\begin{tabular}{lccccccccc}
			\toprule
			 \multirow{2}{*}{ \textbf{Model}} & \multicolumn{3}{c}{ \textbf{Cover-Container}}                        & \multicolumn{3}{c}{ \textbf{Secret-Restored}}                       & \multirow{2}{*}{DR$\downarrow$} & \multirow{2}{*}{Acc$\uparrow$} \\ \cmidrule(lr){2-4}\cmidrule(lr){5-7}
			& MSE$\downarrow$               & PSNR$\uparrow$               & MS-SSIM$\uparrow$            & MSE$\downarrow$               & PSNR$\uparrow$               & MS-SSIM$\uparrow$            &                     &                      \\  \toprule
			 HliPS*            & 0.0162          & 17.88          & 0.9096          & 0.0019          & 27.08          & 0.9646          & 0.480               & 0.8177               \\
			 SteganoGAN*      & 0.0206          & 16.84          & 0.8909          & 0.0089          & 20.46          & 0.9085          & \textbf{0.466}      & 0.8473               \\
			 HiDDeN*           & 0.0123          & 19.11          & 0.8957          & 0.0078          & 21.05          & 0.9727          & 0.490               & 0.8373               \\
			 StegNet*          & 0.0085 & 20.70 & 0.9363          & 0.0056 & 22.47 & 0.9767          & 0.472               & 0.8493               \\
			 Ours & \textbf{0.0063}          & \textbf{22.00}          & \textbf{0.9864} & \textbf{0.0014}          & \textbf{28.53}          & \textbf{0.9962} & 0.479               & \textbf{0.8690}      \\ \toprule
		\end{tabular}
\end{table*}

%% file: tables/deep_steganalysis.tex
\begin{table*}[htbp]
    \caption{The neural steganalysis results. Both the two cases are trained with 4000 image pairs, and tested on 1000 image pairs. In the first case, training data and testing data come from the same specific steganography method. In the second case, the testing data comes from the specific method, but the training data consists of the other four methods evenly.}
    \centering
    \begin{tabular}{lcccccccc}
        \toprule
		\textbf{Steganalyser} & \textbf{Publication} & \textbf{Year}
        & \textbf{Model} & \textbf{HiDDeN} & \textbf{HliPS} & \textbf{SteganoGAN} & \textbf{StegNet} & \textbf{Ours} \\ \midrule
        \multirow{2}{*}{ \textbf{DLHR~\cite{ye2017deep}}} & \multirow{2}{*}{TIFS} & \multirow{2}{*}{2017}
        & First case & 1.000 & 1.000 & 1.000 & 1.000 & 1.000 \\ 
        & & & Second case & 0.889 & 0.782 & 0.949 & 0.929 & 0.891 \\ \midrule
		\multirow{2}{*}{ \textbf{Yedroudj-net~\cite{yedroudj2018yedroudj}}} & \multirow{2}{*}{ICASSP} & \multirow{2}{*}{2018}
		& First case & 1.000 & 1.000 & 1.000 & 1.000 & 1.000 \\ 
		& & & Second case & 0.538 & 0.613 & 0.955 & 0.690 & 0.533 \\       
%

        \bottomrule
    \end{tabular}
    \label{deep_steganalysis}
\end{table*}

%% file: sections/conclusion.tex
\section{Conclusion}
In this paper, we point out the threat of the conventional high-capacity image steganography, and propose a novel Multitask Identity-Aware Image Steganography (MIAIS) framework. We introduce a content loss to preserve the identity information of  secret images, and design a minimax optimization focusing on the discriminability of container images through decreasing the intra-class distances and increasing the inter-class distances. 
Furthermore, we incorporate the restoration module to our MIAIS framework and derive a flexible multitask scenario version, which is adapted to both restoration and recognition. The experiments show the effectiveness of MIAIS compared with the other visual information hiding methods, the robustness of the recognition results across different cover datasets, and the better performance compared to the state-of-the-art high-capacity image steganography methods on face recognition. We believe this work opens up new avenues for privacy preservation based on the visual information hiding methods by achieving direct recognition on container images.

\section*{Acknowledgment}

This work is supported in part by National Key Research and Development Program of China under Grant 2020AAA0107400, Zhejiang Provincial Natural Science Foundation of China under Grant LR19F020004, key scientific technological innovation research project by Ministry of Education, and National Natural Science Foundation of China under Grant U20A20222. This work was supported in part by Ant-Zhejiang University Research Institute of FinTech.
The authors would like to thank Guijie Zhao, Jiaming Ji, Boyu Dong, Xiaoyang Wang for their valuable suggestions. The authors are deeply indebted to anonymous reviewers for their constructive suggestions and helpful comments.

%% file: tables/coeffients.tex
\begin{table}[h]
	\centering
	\caption[TABLE]{Coefficients To Balance The Weight Among The Loss Terms. 'T' represents the hiding threshold (T = 0.02).}
	\label{coeffients}
	\setlength{\tabcolsep}{2mm}
	\resizebox{\textwidth}{14mm}{
	\begin{tabular}{@{}ccccccccccc@{}}
		\toprule
		\multicolumn{1}{c}{\multirow{2}{*}{Scenario}}             &\multicolumn{1}{c}{\multirow{2}{*}{Adjustment}}             & \multicolumn{3}{c}{\textbf{Stage A}}                                                                     & \multicolumn{6}{c}{\textbf{Stage B}}                                                                                                                                                             \\ \cmidrule(lr){3-5}\cmidrule(lr){6-11} 
		\multicolumn{2}{c}{}                                      & \multicolumn{1}{l}{$\mathcal{L}_{id}$($f_{1}$,$h$)} & \multicolumn{1}{l}{$\mathcal{L}_{id}$($f_{2}$,$h$)} & \multicolumn{1}{l}{$\mathcal{L}_{dis}$($f_{1}$,$f_{2}$)} & \multicolumn{1}{l}{$\mathcal{L}_{vis}$($h$)} & \multicolumn{1}{l}{$\mathcal{L}_{id}$($f_{1}$,$h$)} & \multicolumn{1}{l}{$\mathcal{L}_{id}$($f_{2}$,$h$)} & \multicolumn{1}{l}{$\mathcal{L}_{cont}$($h$)} & \multicolumn{1}{l}{$\mathcal{L}_{dis}$($f_{1}$,$f_{2}$)} & \multicolumn{1}{l}{$\mathcal{L}_{res}$($h$,$r$)} \\ \toprule
		\multirow{2}{*}{Direct Recognition} & $\mathcal{L}_{vis}$\textgreater{}T & 5                             & 5                             & -0.5                            & 10                          & 0.01                          & 0.01                          & 1                            & 20                              & -                        \\
		& $\mathcal{L}_{vis}$$\leq$T   & 10                            & 10                            & -3                              & 3                           & 0.01                          & 0.01                          & 1                            & 4                               & -                        \\ \toprule
		\multirow{2}{*}{Multi-task}         & $\mathcal{L}_{vis}$\textgreater{}T & 15                            & 15                            & -1.5                            & 5                           & 0.01                          & 0.01                          & 1                            & 15                              & 3.75                     \\
		& $\mathcal{L}_{vis}$$\leq$T   & 15                            & 15                            & -1.5                            & 1                           & 0.01                          & 0.01                          & 1                            & 15                              & 1                        \\ \toprule
	\end{tabular}
}
\end{table}

%% file: tables/robust_cover_high_PSNR.tex
\begin{table*}[h]
	\caption{Robustness on Different Cover Sets with AFD Secret Set.
		$\uparrow$ means the higher the better; $\downarrow$ means the lower the better. ``100-SS'' denotes to the original 100 cover set form style transfer and satellite datasets.
	}
	\centering
	\label{Table_cover_robustness_high_PSNR} 
	\begin{tabular}{lccccc}
		\toprule
		{\textbf{Cover Set}} & MSE$\downarrow$ & PSNR$\uparrow$ & MS-SSIM$\uparrow$ & Acc$\uparrow$ & Original Acc$\uparrow$ \\
		\toprule
		100-SS &	0.0011 & 29.67 & 0.9935  & 0.8585 & 0.8752 \\
		LFW & 0.0010 & 30.07  & 0.9946  & 0.8495 & 0.8730 \\
		PASCAL-VOC12  & 0.0008 & 31.19  & 0.9953  & 0.8578 & 0.8724 \\
		ImageNet &  0.0011 & 29.65  & 0.9944  & 0.8626 & 0.8729 \\
		\bottomrule
	\end{tabular}
\end{table*}

%% file: tables/comp_steg_other_high_PSNR.tex
\begin{table*}[h]
	\centering
	\caption[TABLE]{Comparison with State-of-the-Art High-Capacity Image Steganography Methods on the AFD Dataset.
	$\uparrow$ means the higher the better; $\downarrow$ means the lower the better.
	}
		\label{Table_Comparison_Steganography_high_PSNR} 
		\begin{tabular}{lcccccccc}
			\toprule
			 \multirow{2}{*}{ \textbf{Model}} & \multicolumn{3}{c}{ \textbf{Cover-Container}}                        & \multicolumn{3}{c}{ \textbf{Secret-Restored}}                       & \multirow{2}{*}{Acc$\uparrow$}& \multirow{2}{*}{Original Acc$\uparrow$} \\ \cmidrule(lr){2-4}\cmidrule(lr){5-7}
			& MSE$\downarrow$               & PSNR$\uparrow$               & MS-SSIM$\uparrow$            & MSE$\downarrow$               & PSNR$\uparrow$               & MS-SSIM$\uparrow$            &                     &                      \\  \toprule
			 HliPS* & 0.0012 & 29.35 & 0.9687 & 0.0014 & 28.54 & 0.9808  & 0.6195 & 0.8177 \\
			 SteganoGAN* & 0.0114 & 19.42 & 0.9484 & 0.0449 & 13.47 & 0.8665  & 0.5282 & 0.8473 \\
			 HiDDeN* & 0.0011 & 29.59 & \textbf{0.9850} & 0.0070 & 21.55 & 0.9239  & 0.6705 & 0.8373 \\
			 StegNet* & 0.0019 & 27.28 & 0.9733 & 0.0066 & 21.78 & 0.9524 & 0.4990 & 0.8493 \\
			 Ours & \textbf{0.0010} & \textbf{30.09} & 0.9849 & \textbf{0.0007} & \textbf{31.37} & \textbf{0.9811} & \textbf{0.7955} & 0.8690 \\ \toprule
		\end{tabular}
\end{table*}

%% file: tables/tiny_dataset_with_restorer.tex
\begin{table*}[h]
	\centering
	\caption[TABLE]{Comparison with State-of-the-Art High-Capacity Image Steganography Methods on the tiny-AFD Dataset. $\uparrow$ means the higher the better; $\downarrow$ means the lower the better. The recognition accuracy (Acc) in the secret image is 0.7379.}
		\label{tinyset} 
		\begin{tabular}{lccccccccc}
			\toprule
			 \multirow{2}{*}{ \textbf{Model}} & \multicolumn{3}{c}{ \textbf{Cover-Container}} & \multicolumn{3}{c}{ \textbf{Secret-Restored}} & \multirow{2}{*}{DR$\downarrow$} & \multirow{2}{*}{Acc$\uparrow$}\\ 
			 \cmidrule(lr){2-4}\cmidrule(lr){5-7}
			& MSE$\downarrow$               & PSNR$\uparrow$               & MS-SSIM$\uparrow$            & MSE$\downarrow$               & PSNR$\uparrow$               & MS-SSIM$\uparrow$            &                     &                      \\  \toprule
			 
			 HliPS* & 0.0015 & 28.36 & 0.9634 & 0.0026 & 25.88 & 0.9632 & 0.481 & 0.6090  \\
			 SteganoGAN* & 0.0123 & 19.09 & 0.9349 & 0.0184 & 17.36 & 0.9355 & 0.475 & 0.5162 \\
			 HiDDeN* & 0.0016 & 27.96 & 0.9771 & 0.069 & 21.61 & 0.9400 & 0.459 & 0.6522\\
			 StegNet* & 0.0024 & 26.20 & 0.9691 & 0.0070 & 21.55 & 0.9469& 0.474 & 0.4855 \\
			 Ours & 0.0012 & 29.21 & 0.9789 & 0.0005 & 33.01 & 0.9865 & 0.461& 0.7190\\
			 \toprule
		\end{tabular}
\end{table*}